\documentclass[10pt,twocolumn,letterpaper]{article}

\usepackage{cvpr}

\usepackage{epsfig}
\usepackage{graphicx}
\usepackage[small]{caption}

\usepackage{amsmath}
\usepackage{amssymb}

\usepackage[american]{babel}
\usepackage{arydshln}
\usepackage{enumitem}
\usepackage[normalem]{ulem}

\usepackage{array}
\usepackage{booktabs} 
\usepackage{multirow} 

\usepackage{times}
\usepackage{color}
\usepackage{pifont} 

\usepackage{url}  
\usepackage[pagebackref=true,breaklinks=true,letterpaper=true,colorlinks,citecolor =blue, bookmarks=false]{hyperref}

\usepackage[table]{xcolor}

\usepackage{rotating}

\usepackage{soul}
\usepackage[utf8]{inputenc}

\usepackage[linesnumbered,ruled]{algorithm2e}

\usepackage{bm}

\usepackage{siunitx}
\definecolor{dgreen}{rgb}{0, 0.6, 0} 
\definecolor{cyan}{rgb}{0, 0.5, 0.6} 
\definecolor{Darkviolet}{rgb}{0.58, 0, 0.83}

\DeclareMathOperator*{\argmin}{arg\,min}

\newcommand{\heading}[1]{\noindent\textbf{#1}}

\newcommand{\figref}[1]{Figure~\ref{fig:#1}}
\newcommand{\tabref}[1]{Table~\ref{tab:#1}} 
\newcommand{\eqnref}[1]{Equation~(\ref{eq:#1})}

\newcommand{\algoref}[1]{Algorithm~\ref{algo:#1}}

\newcommand{\ignore}[1]{}   

\makeatletter
\def\hlinewd#1{%
  \noalign{\ifnum0=`}\fi\hrule \@height #1 \futurelet
   \reserved@a\@xhline}
\makeatother

\newcolumntype{L}[1]{>{\raggedright\arraybackslash}p{#1}}
\newcolumntype{C}[1]{>{\centering\arraybackslash}p{#1}}
\newcolumntype{R}[1]{>{\raggedleft\arraybackslash}p{#1}}

\newcommand{\cmark}{\checkmark}
\newcommand{\xmark}{$\times$}

\graphicspath{{figure/}}

\def\inImg{{\bm x}}
\def\recImg{\hat{\bm x}}
\def\latentZ{{\bm z}}
\def\spFeat{{\bm f}}
\def\personSym{\mathcal{P}}
\def\lightSym{\mathcal{I}}

\setlist[itemize]{leftmargin=*}
\newcommand{\erhao}{\fontsize{21pt}{\baselineskip}\selectfont}

\cvprfinalcopy 

\def\cvprPaperID{xxxx} 

\ifcvprfinal\fi

\begin{document}

{\onecolumn

\noindent \textbf{\erhao{Illumination-Adaptive Person Re-identification}}

\vspace{2cm}

\noindent {\large{Zelong Zeng, Zhixiang Wang, Zheng Wang, Yinqiang Zheng, Yung-Yu Chuang, Shin’ichi Satoh}}

\vspace{2cm}


\vspace{1cm}

\noindent For reference of this work, please cite:

\vspace{1cm}
\noindent Zelong Zeng, Zhixiang Wang, Zheng Wang, Yinqiang Zheng, Yung-Yu Chuang, Shin’ichi Satoh.
``Illumination-Adaptive Person Re-identification.''
In \emph{IEEE Transactions on Multimedia.} 2020.

\vspace{1cm}

\noindent Bib:\\
\noindent
@article\{IAREID-tmm2020,\\
\ \ \  title=\{Illumination-Adaptive Person Re-identification\},\\
\ \ \  author=\{Zelong Zeng, Zhixiang Wang, Zheng Wang, Yinqiang Zheng, Yung-Yu Chuang, Shin’ichi Satoh\},\\
\ \ \  journal=\{IEEE Transactions on Multimedia\},\\
\ \ \  year=\{2020\},\\
\ \ \  volume=\{\},\\
\ \ \  number=\{\},\\
\ \ \  pages=\{\}\\
\}
}

\twocolumn

\title{Illumination-Adaptive Person Re-identification}

 \author{Zelong~Zeng$^{1}$\thanks{Co-first Authors}
    \hspace{0.1in} Zhixiang~Wang$^{2 *}$
    \hspace{0.1in} Zheng~Wang$^{3}$\thanks{Corresponding Author}
    \hspace{0.1in} Yinqiang Zheng$^{3}$\\
	\hspace{0.2in} Yung-Yu Chuang$^{2}$
	\hspace{0.2in} Shin'ichi Satoh$^{1,3}$
	\vspace{1mm} \\
    $^{1}$The University of Tokyo \hspace{0.1in} 
    $^{2}$National Taiwan University \hspace{0.1in} $^{3}$National Institute of Informatics\\
    \texttt{\small{\{zzlbz, wangz, satoh\}@nii.ac.jp, \{r06944046, cyy\}@csie.ntu.edu.tw}}
}

\maketitle
\thispagestyle{empty}

\begin{abstract}
Most person re-identification (ReID) approaches assume that person images are captured under relatively similar illumination conditions. In reality, long-term person retrieval is common, and person images are often captured under different illumination conditions at different times across a day. In this situation, the performances of existing ReID models often degrade dramatically. This paper addresses the ReID problem with illumination variations and names it as {\em Illumination-Adaptive Person Re-identification (IA-ReID)}. We propose an Illumination-Identity Disentanglement (IID) network to dispel different scales of illuminations away while preserving individuals' identity information. To demonstrate the illumination issue and to evaluate our model, we construct two large-scale simulated datasets with a wide range of illumination variations. Experimental results on the simulated datasets and real-world images demonstrate the effectiveness of the proposed framework.
\end{abstract}


\section{Introduction}
\label{sec:intro}

Person re-identification (ReID) is a cross-camera retrieval task. Given a query person-of-interest, it aims to retrieve the same person from a database of images collected from multiple cameras~\cite{wang2016zero,ye2016person,wang2016deeplist,wang2018person,wang2019incremental,zhong2018camera,wang2019learning}. The key challenge of ReID lies in the person's appearance variations among different cameras. Most previous methods attempt to find a feature representation that is stable to the appearance variations. They have well investigated how to deal with variations in occlusions~\cite{zheng2015partial}, resolutions~\cite{wang2016scale,wang2018cascaded}, poses~\cite{ge2018fd}, \etc. However, the influence of ever-changing illumination conditions has been largely ignored. Most popular ReID datasets, such as Market1501~\cite{zheng2015scalable} and DukeMTMC-reID~\cite{zheng2017unlabeled}, have relatively uniform illumination conditions, as their images were captured under similar illumination at the same period of time.  

\tabcolsep=2pt
\begin{figure}[t]
\centering 
\renewcommand{\arraystretch}{1.25}
\begin{tabular}{c}
\includegraphics[width=0.85\columnwidth]{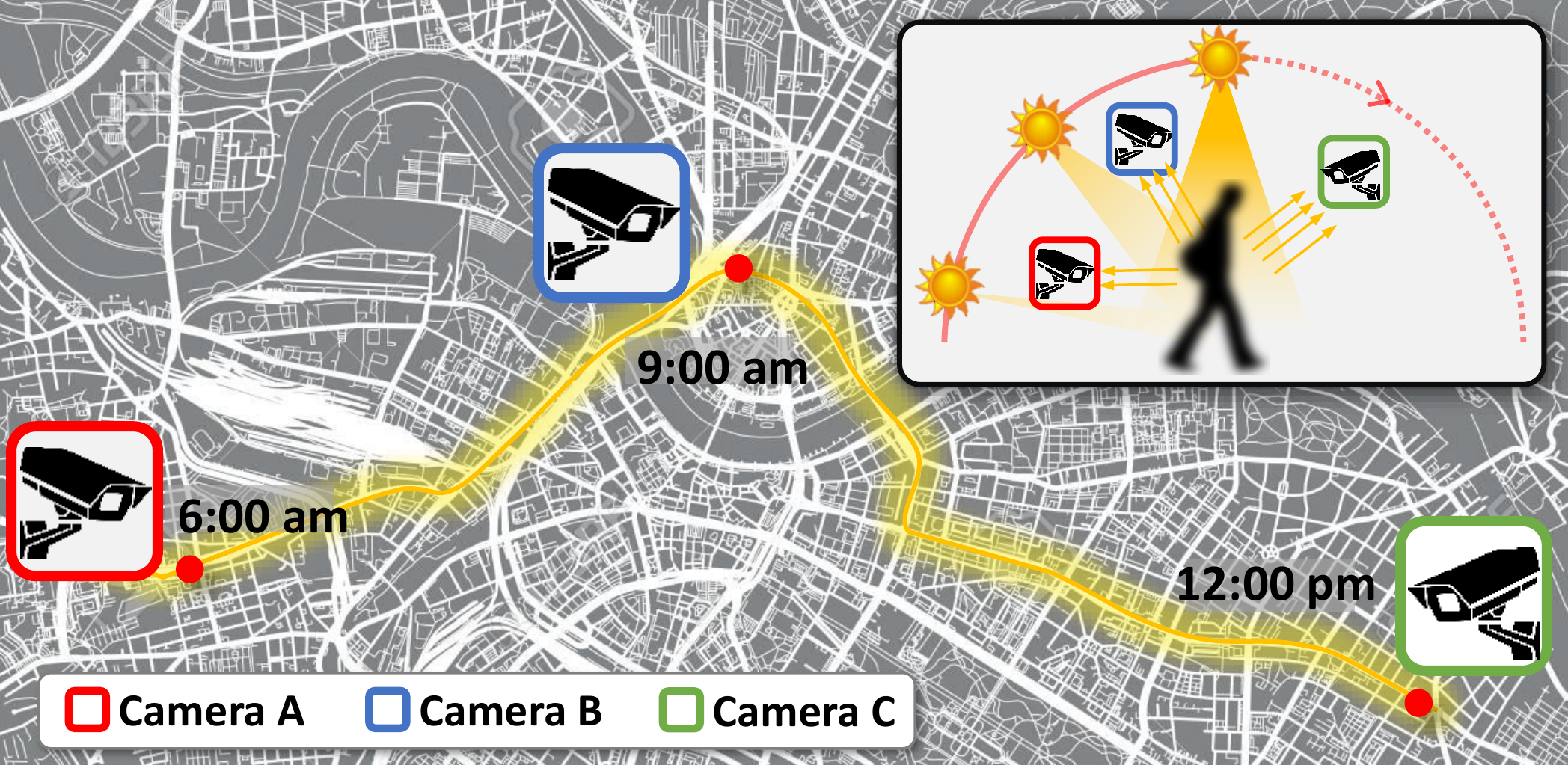}\\
\small{(a)}\\
\includegraphics[width=0.85\columnwidth]{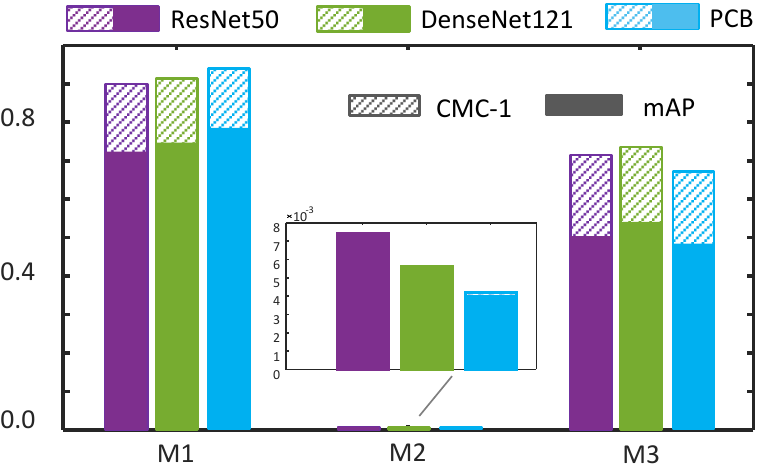} \\ %
\small{(b)}\\
\end{tabular}
\caption{(a) A toy example to illustrate the real application scenario of ReID where images captured at different times could have quite different illumination conditions. (b) The results of preliminary experiments to show the impact of illumination-adaptive. The experiments were conducted on three kinds of networks, including ResNet50 [He \emph{et al.}, 2016], DenseNet121 [Huang \emph{et al.}, 2017], and PCB [Sun \emph{et al.}, 2018]. ``M1'', ``M2'' and ``M3'' stand for the CMC-1 and mAP results for three different training and testing pairs. ``M1'' is obtained by training and testing on the Market-1501 dataset. ``M2'' is obtained by training on the Market-1501 and testing on the Market-1501++ dataset. ``M3'' is obtained by training and testing on the Market-1501++ dataset.}
\label{fig:motiv}
\end{figure}

In practice, long-term person retrieval is often required in video surveillance networks and criminal investigation applications. As \figref{motiv}(a) shows, the images of a suspect could be taken under very different illumination conditions at different times across a day. He may appear in camera $A$ with dim light at 6:00~a.m., then in camera $B$ with normal light at 9:00~a.m., and finally in camera $C$ with glare light at 12:00~p.m. Existing researches do not investigate this illumination-adaptive issue. We name the ReID task under different illumination conditions as {\em Illumination-Adaptive Person Re-identification (IA-ReID)}. In this task, given a probe image under one illumination, the goal is to match gallery images with several different illuminations. The images can be with normal illumination as existing ReID datasets (Market1501, DukeMTMC-reID), can be very bright if captured under dazzling sunshine, and also can be very dark if captured in the sunset or even during the night.

The illumination greatly affects the performance of re-identification. As we know, the re-identification performance heavily relies on the characteristics of the datasets. The model trained on one dataset often can not perform well on the other. Traditional models, although efficient and effective to re-identify gallery images with the same illumination, may suffer from a significant performance drop when the illumination conditions of gallery images vary greatly. We have conducted preliminary experiments for demonstrating the performance degradation using the Market-1501 dataset. We simulated different illumination conditions for images in the dataset. The resultant dataset with varying illumination conditions is named the Market-1501++ dataset. Three training-testing configurations were performed: M1 (both train and test on Market-1501), M2 (train on Market-1501 but test on Market-1501++), and M3 (both train and test on Market-1501++). As \figref{motiv}(b) illustrates, 1) the learned models are not stable across datasets with different illuminations (``M2'') and 2) even trained with images under different illumination conditions, the model cannot achieve satisfying performance (``M3''). That is to say, general ReID models lose their effectiveness in the situation with illumination variations. 

As far as we know, some researches investigated the issue of different illuminations in ReID~\cite{bhuiyan2015exploiting,wang2014camera}. However, they only consider a situation of two scales of illuminations. They assumed that the probe and the gallery images are respectively captured from two cameras, each with a corresponding illumination condition. In such a controlled setting, they proposed to learn the relationship between two scales of illuminations by brightness transfer functions~\cite{bhuiyan2015exploiting} or the feature projection matrix~\cite{wang2014camera}. IA-ReID is a more practical problem with multiple illuminations. Obviously, constructing the relationships among different scales of illuminations is not a practical solution for this problem. If there are ten different scales in the dataset, the method needs to construct ten different relationships, and it cannot be guaranteed the ten relationships work perfectly.

Removing the effect of illumination is another intuitive idea. One solution is to do image enhancement~\cite{chen2018learning} for the low-illumination images and image reconstruction~\cite{eilertsen2017hdr} for the images with high exposure. However, this kind of methods either cannot handle extreme illuminations~\cite{chen2018deep}, or are particularly designed for visualization and rely heavily on the data condition and training samples~\cite{chen2018learning,eilertsen2017hdr}.

Another solution is to disentangle the illumination information from the person feature. This idea is learned from the existing face recognition methods~\cite{tran2017disentangled}, where certain face attributes are separated from the feature vector. In this paper, we follow this thread of ideas and propose an Illumination-Identity Disentanglement (IID) network. Inspired by previous researches dealing with the image resolution issue~\cite{wang2018cascaded}, we construct two simulated illumination-adaptive datasets based on Market1501~\cite{zheng2015scalable} and DukeMTMC-reID~\cite{zheng2017unlabeled}. Then, these two datasets are utilized to evaluate the effectiveness of the IID network. Our contributions can be summarized as follows:
\begin{itemize}
\item {\bf A new and practical problem.} We raise a new and practical task, \ie, IA-ReID. The task is practical for long-term person re-identification applications. We construct two simulated datasets with different illuminations to put forward this task. Most general ReID models are proved ineffective when evaluated on these two datasets.
\item {\bf A novel method.} We propose a novel Illumination-Identity Disentanglement (IID) network, which dispels the illumination information away from a person's appearance. The method achieves great performance improvement on our two datasets. We also evaluate our model on some real images, and it is capable of alleviating the effect of illumination discrepancy.
\item {\bf Simplicity and Effectiveness.} We construct the IID network based on a simple backbone. The network is easy to follow. Extensive experiments prove that IID is robust in long-term person re-identification applications. In this way, we set a benchmark for the new task. 
\end{itemize}

\section{Related work}

\subsection{Short-term ReID}
In the short-term condition, existing researches paid attention to the challenges from resolution variations~\cite{wang2018cascaded}, pose changes~\cite{ge2018fd} and occlusion~\cite{zheng2015partial}. Many methods were proposed to learn robust representations to overcome those challenges. They have achieved very high performances on the public datasets. Liao \etal~\cite{liao2015person} analyzed the horizontal occurrence of local features and maximized the occurrence to make a stable representation against viewpoint changes. Su \etal~\cite{su2017pose} leveraged the human part cues to alleviate the pose variations and learn robust feature representations from both the global image and different local parts. Zheng \etal~\cite{zheng2015partial} fused a local patch-level and a global part-based matching model to address the occlusion problem. However, when the re-identification task goes to the long-term situation, the illumination variations come to be the key issue. 

We have claimed that public datasets, such as Market1501~\cite{zheng2015scalable} and DukeMTMC-reID~\cite{zheng2017unlabeled}, have relatively uniform illumination conditions. As evidence, for both Market1501 and DukeMTMC-reID, the file name of each image includes the frame index, which indicates the relative captured time of the corresponding image. The range of frame indexes of all images spans over 90,000 frames for Market1501, i.e., those images were captured in a time span of around one hour (25fps$\times$3600s). For DukeMTMC-reID, images were captured in around two hours. It supports our claim that existing ReID datasets were captured in a short period of time, and the images have relatively uniform illumination conditions.

\begin{figure*}[ht] 
\centering
\includegraphics[width=0.9\textwidth]{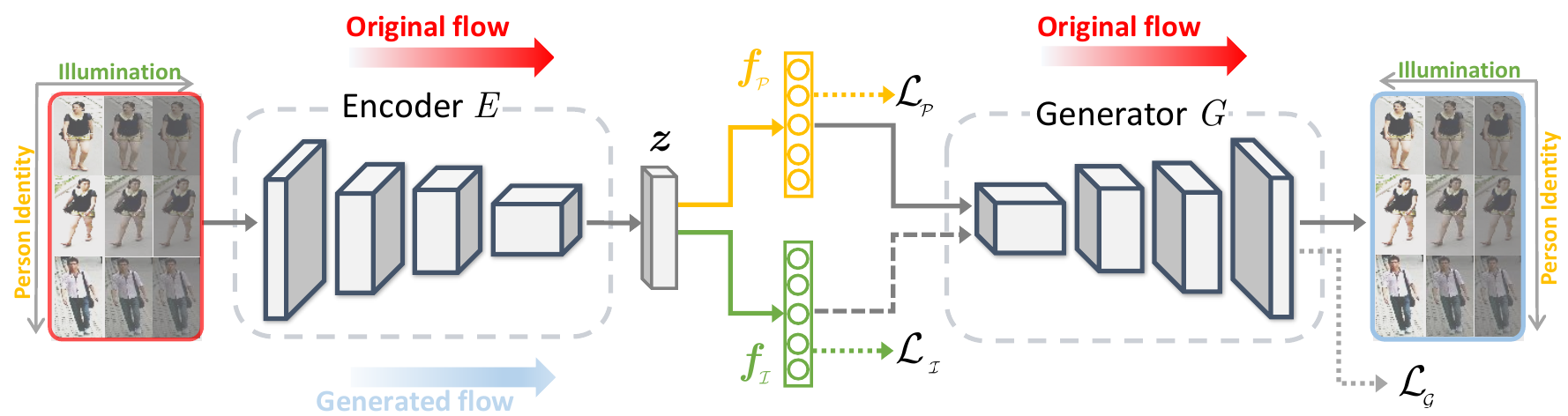}
\caption{{\bf The architecture of our proposed Illumination-Identity Disentanglement (IID) network.} It consists of an encoder $E$, a generator $G$, and two feature embedding layers $\mathcal{H}_\personSym$ and $\mathcal{H}_\lightSym$. The network is optimized through three loss functions. The embedded person (identity-relevant) feature $\spFeat_\personSym$ is trained with the ReID loss $\mathcal{L}_{\personSym}$; the embedded identity-irrelevant feature $\spFeat_\lightSym$ is trained with the illumination loss $\mathcal{L}_{\lightSym}$ and the generation loss $\mathcal{L}_\mathcal{G}$ is used to guide the optimization of the generator. The training process has two work flows. 1) The original flow is defined as $\inImg \!\xrightarrow[]{E}\! \latentZ \!\xrightarrow{\mathcal{H}}\! (\spFeat_\personSym, \spFeat_\lightSym) \!\xrightarrow{G}\! \recImg$. The encoder $E$ firstly encodes an image into a latent vector $\latentZ$, and then disentangles into two components $\spFeat_\personSym$ and $\spFeat_\lightSym$. The generator $G$ reconstructs the image $\recImg$ fed with the combination of $\spFeat_\personSym$ and $\spFeat_\lightSym$. 2) The generated flow is defined as $\recImg \!\xrightarrow[]{E}\! \hat\latentZ \!\xrightarrow{\mathcal{H}}\! (\hat\spFeat_\personSym, \hat\spFeat_\lightSym)$. To enhance the identity-preserving of the reconstructed image $\recImg$, we feed it into the network again. Similarly, we extract the disentangled representations $\hat\spFeat_\personSym$ and $\hat\spFeat_\lightSym$.}
\label{fig:framework}
\end{figure*}

\subsection{Illumination Problem in ReID}
Some researches start to investigate the illumination issue in ReID. Previous methods~\cite{kviatkovsky2013color,bhuiyan2015exploiting,wang2014camera} consider the situation that the probe and the gallery images respectively captured from two cameras with two different illuminations. Kviatkovsky \etal~\cite{kviatkovsky2013color} proposed an invariant feature exploiting a structure of color distributions, using different parts of the person. Bhuiyan \etal~\cite{bhuiyan2015exploiting} learned robust brightness transfer functions to release the illumination change from one camera to the other. Wang \etal~\cite{wang2014camera} designed a feature projection matrix to project image features of one camera to the feature space of another camera. Ma \etal~\cite{ma2018low} focused on the low illumination problem. They transformed all the images to a uniform low illumination and proposed metric learning methods to address the low illumination. There are also some researches related to more extreme illumination conditions. They considered that when the imaging condition goes to late at night, the low-illumination image will transform into the infrared image. To this end, some methods~\cite{wu2017rgb,ijcai18vtreid,dai2018cross,ye2019bi,wang2019duallevel,wang2019beyond,kansal2020sdl} paid their attention to the infrared-visible re-identification, where one part of the images are from the visible camera and the other part is from the infrared camera. They designed networks to bridge the gap between infrared and visible images. In this paper, multiple illuminations are taken into consideration, including not only the low illumination images but also the high illumination ones. Our illumination-adaptive setting is more practical than previous methods for long-term ReID.

\subsection{Disentangled Feature Learning}
Some previous works tried to disentangle the representations in different kinds of recognition tasks~\cite{yan2014efficient,yan2018fast,yan2019cross,yan2019stat}, for example pose-invariant recognition~\cite{tran2017disentangled} and identity-preserving image editing~\cite{li2016deep}. They exploited attribute supervision and encoded each attribute as a separate element in the feature vector. Liu \etal~\cite{liu2018exploring} proposed to learn disentangled but complementary face features with face identification. Disentangled feature learning was investigated in image-to-image translation tasks as well. Lee \etal~\cite{lee2018diverse} exploited disentangled representation for producing diverse outputs, in particular, a domain-invariant content space capturing shared information across domains and a domain-specific attribute space was proposed. Gonzalez \etal~\cite{gonzalez2018image} introduced the concept of cross-domain disentanglement and separated the internal representation into one shared part and two exclusive parts. There also exist some methods using disentangled representations to address the single-image deblurring tasks~\cite{lu2019unsupervised}. As far as we know, there is no method designed to disentangle features from individuals in ReID. Our method is the first to consider disentangling the illumination, one kind of identity-irrelevant information.

\subsection{Encoder-Decoder Network}
Our method exploits an encoder-decoder network. As we know, this framework is prevalent, \eg, the Ladder Network~\cite{rasmus2015semi} and U-Net~\cite{ronneberger2015u}. Although our approach is similar to these two typical networks, there are key differences. 1) Different focuses: The Ladder Network focuses on the semi-supervised learning task, the U-Net focuses on the image segmentation task, while our method focuses on disentangling the person and illumination features. 2) Different structures: The Ladder Network consists of a supervised part and an unsupervised part, the U-net has no ID supervision and pays its attention to image generation, while our method introduces two supervised parts to extract two kinds of features. 3) Different outputs: The Ladder Network attempts to output a label for each input, the U-Net tries to obtain an image output, while our method attempts to extract features.

\section{Our method}
\label{sec:method}

\subsection{Overview}
\figref{framework} depicts the overall architecture of our Illumination-Identity Disentanglement (IID) network. The encoder $E$ takes ResNet-50 as the backbone network and encodes the input image $\inImg$ into a latent vector $\latentZ \!=\! E(\inImg)$, where ${\bm z}\!\in\!\mathbb{R}^{2048}$. Next, two independent fully-connected (\emph{FC}) layers $\mathcal{H}_\personSym$ and $\mathcal{H}_\lightSym$, are employed to project the latent $\latentZ$ into two different feature vectors, \ie, the person feature $\spFeat_\personSym \!=\! \mathcal{H}_\personSym(\latentZ)$ and the identity-irrelevant feature $\spFeat_\lightSym \!=\! \mathcal{H}_\lightSym(\latentZ)$. Note that $\spFeat_\personSym$ and $\spFeat_\lightSym$ are expected to be respectively stable to illumination variations and irrelevant to person identity. To enforce the disentangled information fully represent the input image, the generator $G$ is used to reconstruct the image $\recImg \!=\! G(\spFeat_\personSym, \spFeat_\lightSym)$ to approximate the input image $\inImg$ from the disentangled feature vectors, $\spFeat_\personSym$ and $\spFeat_\lightSym$. Here, $\spFeat_\personSym$ and $\spFeat_\lightSym$ are concatenated, and act as the input of the generator $G$. The reason why we maintain the disentangled illumination-relevant feature rather than simply getting rid of it is to reduce the potential loss of discriminative information in the identity-relevant feature. Specifically, we use the generator $G$ and feed it with the disentangled identity-relevant and identity-irrelevant features to reconstruct the input with minimum loss of information. If we dump the disentangled illumination-relevant feature without guaranteeing the reconstruction, some useful information could slip away with the removed illumination-relevant feature.


\subsection{Disentangled Feature Learning}

\heading{Identity-relevant feature learning.} Given the encoded latent vector $\latentZ$ from the encoder $E$, the \emph{FC} layer $\mathcal{H}_\personSym$ projects it to the person feature $\spFeat_\personSym$, where $\spFeat_\personSym \!=\! \mathcal{H}_\personSym(\latentZ)$. Since the feature $\spFeat_\personSym$ is required to capture information relevant to person identity, we use the ReID loss $\mathcal{L}_{\personSym}$ and person identity information to supervise the training process. 
The ReID loss $\mathcal{L}_{\personSym}$ combines the triplet loss $\mathcal{L}_{\personSym}^T$ and the softmax loss $\mathcal{L}_{\personSym}^S$ and it can be written as 
\begin{equation}
\label{eq:loss_person}
\mathcal{L}_{\personSym} = \lambda_1\mathcal{L}_{\personSym}^T+\lambda_2\mathcal{L}_{\personSym}^S\,,
\end{equation}
where the weights $\lambda_1$ and $\lambda_2$ 
are used for balancing these two losses. The training strategy is the same as the one in the general ReID framework; hence the extracted feature can be used for the identification task directly.

The triplet loss is used for similarity learning and it can be formulated as
\begin{equation}
\label{eq:tlloss_person}
\mathcal{L}_{\personSym}^T = \sum_{\spFeat_{\personSym}^a, \spFeat_{\personSym}^p, \spFeat_{\personSym}^n \!\in\! \mathcal{B}}\left[\mathcal{D}(\spFeat_{\personSym}^a, \spFeat_{\personSym}^p) - \mathcal{D}(\spFeat_{\personSym}^a, \spFeat_{\personSym}^n) + \xi\right]_{+}\,,\\
\end{equation}
where $\mathcal{B}$ represents a mini-batch consisting of extracted person features $\spFeat_{\personSym}$. For an anchor feature vector $\spFeat_{\personSym}^a$, the positive sample $\spFeat_{\personSym}^p$ and the negative sample $\spFeat_{\personSym}^n$ respectively denotes a feature vector having the same identity with $\spFeat_{\personSym}^a$ and one with different identity from $\spFeat_{\personSym}^a$. Note that $\spFeat_{\personSym}^a \!\ne\! \spFeat_{\personSym}^p$. $\xi$ is a margin parameter; $\mathcal{D}(\boldsymbol{\cdot})$ calculates the Euclidean distance; and $[\boldsymbol d ]_{+}=\max(d, 0)$ truncates negative numbers to zero while keeping positive numbers the same. Note that we exploit the most primitive triplet loss function. While during the training process, we select hard samples for each triplet. For each anchor, we randomly select 16 samples with the same ID and 64 samples with different IDs, and compute the feature distance between the anchor and each selected sample. Then, we select the farthest positive sample as the hard positive sample, and the nearest negative sample as the hard negative sample. They together construct a triplet.

The softmax loss is employed for identity information learning, which is written as
\begin{equation}
\label{eq:crossloss_person}
\mathcal{L}_{\personSym}^S = - \frac{1}{N} \sum _{j=1}^{N} \log {\hat{\bm y}}_\personSym^{j}\,,
\end{equation}
where $N$ is the number of images in the mini-batch $\mathcal{B}$ and $\hat{{\bm y}}_\personSym$ is the predicted probability of the input belonging to the ground-truth class with ${\bm y}_\personSym\!=\!\texttt{softmax}({\bf W}_\personSym \spFeat_\personSym\!+\!{\bf b}_\personSym)$, where ${\bf W}_\personSym$ and ${\bf b}_\personSym$ are the trainable weight and bias of $\mathcal{H}_\personSym$ respectively.

\heading{Identity-irrelevant feature learning.} Given the encoded latent vector $\latentZ$ from the encoder $E$, the \emph{FC} layer $\mathcal{H}_\lightSym$ projects it to the identity-irrelevant feature $\spFeat_\lightSym$, written as $\spFeat_\lightSym \!=\! \mathcal{H}_\lightSym(\latentZ)$. To make the feature irrelevant to person identity, we need to feed the network with images taken under different illuminations. Thanks to our simulated dataset, each image is automatically assigned an illumination label, indicating the scale of relative illumination change. Note that the detail information of the datasets is described in Section~\ref{subsec:dataset}. The illumination label is somehow coarse since we assumed that images in original datasets are captured in a relatively uniform illumination condition. To eliminate the reliance on this assumption, we make two necessary modifications. 1) We adopt the classifier problem to do regression. 2) We use a soft label strategy instead of a hard label. The soft label strategy is used because we would like to leave some room for tolerating slight changes, possibly caused by camera styles and viewpoints, occurring on this relatively uniform illumination condition. For the same purpose, we also transform the classification problem into a regression problem.

The regression loss is written as
\begin{equation}
\label{eq:crossloss_light}
\mathcal{L}_{\lightSym} = \frac{1}{N} \sum _{j=1}^{N}\left\|{\hat c}_\lightSym^j -({\bf W}_\lightSym \spFeat_\lightSym^j\!+\!{\bf b}_\lightSym)\right\|_2^2\,,
\end{equation}
where ${\bf W}_\lightSym$ and ${\bf b}_\lightSym$ are the trainable weight and bias respectively. The soft label ${\hat c}_\lightSym$ is the summation of the ground truth label $c_\lightSym$ and Gaussian noise $\epsilon$.

\begin{equation}
\label{eq:soft_label}
{\hat c}_\lightSym \!=\! c_\lightSym + \epsilon\,, \ \ \text{with} \ \ \epsilon\in \mathcal{N}(0, 1)\,. 
\end{equation}

Note that there is no difference between the datasets of compared state-of-the-art ReID models and our network. Images with multiple illuminations are just used to separate the illumination information from the person feature.

\subsection{Identity-preserving Image Generation}
The image generator $G$ is employed to ensure that the disentangled information has minimum information loss. It is fed with the combination of $\spFeat_\personSym$ and $\spFeat_\lightSym$ and generates the reconstructed image $\recImg \!=\! G(\spFeat_\personSym, \spFeat_\lightSym)$. We use the \emph{MSE} loss as the supervision information, which is defined as 
\begin{equation}
\label{eq:loss_generator}
\mathcal{L}_\mathcal{G} = \left\|\inImg - \recImg\right\|_2^2\,.
\end{equation}
It is worth mentioning that, in addition to guiding the reconstruction of the input and supervising the training process, $G$ can be used to synthesize images with same identity but varying illuminations by altering the illumination-relevant feature vector $\spFeat_\lightSym$. For the network configure of generator $G$, we use six decode modules. Each decode module consists of a ReLU layer, a 2D transposed convolution layer, a batch normalization layer and a dropout layer. We concatenate $\spFeat_\personSym$ and $\spFeat_\lightSym$ as the input of the generator, and the output of generator is the generated image with size $128\!\times\!64\!\times\!3$.

To enforce the reconstructed image $\recImg$ identity-preserving, we feed it into the network again. As \figref{framework} shows, we name this process as generated flow, in contrast to the original flow. In the generated flow, the loss $\mathcal{L}_\personSym$ and $\mathcal{L}_\lightSym$ are taken into account in the total loss function. However, the features $\spFeat_\personSym$ and $\spFeat_\lightSym$ are not needed to utilize again to generate new images.

\subsection{Training Process}
The training process has three phases: feature disentanglement training, generator training, and joint training. Their details will be described as follows.

\heading{Phase I: Feature disentanglement training.} 
In this phase, parameters of the disentangled feature learning module are updated. As mentioned above, the disentangled feature learning module consists of the encoder $E$, two feature embedding layers $\mathcal{H}_{\personSym}$ and $\mathcal{H}_{\lightSym}$, and the weight and bias ${\mathbf W}_{\personSym/\lightSym}$, ${\mathbf b}_{\personSym/\lightSym}$ for loss functions defined in \eqnref{crossloss_person} and \eqnref{crossloss_light}. We denote the parameters of $E$ by $\theta_E$. The parameters of the FC layer $\mathcal{H}_{\personSym}$, the weights ${\mathbf W}_{\personSym}$ and the biases ${\mathbf b}_{\personSym}$ are denoted by $\theta_\personSym$. Similarly, parameters of $\mathcal{H}_{\lightSym}$, ${\mathbf W}_{\lightSym}$ and ${\mathbf b}_{\lightSym}$ are denoted by $\theta_\lightSym$. The object function for this phase is  
\begin{equation}
\label{eq:fd_obj}
\argmin_{\theta_{E}, \theta_\mathcal{P}, \theta_\mathcal{I}} \mathcal{L}_\personSym + \lambda_3\mathcal{L}_\lightSym\,.
\end{equation}
We set $\lambda_3=1$. Note that $\mathcal{L}_\personSym$ is defined in \eqnref{loss_person} with the hyperparameters $\lambda_1\!=\!\lambda_2\!=\!0.5$. The hyperparameter $\xi$ in \eqnref{tlloss_person} is set to 0.3. 

The encoder $E$ is initialized with ImageNet pre-trained weights~\cite{deng2009imagenet}. Other parts are initialized with He's method \cite{he2015delving}. The optimizer utilizes SGD with momentum and weight decay set to 0.9 and \num{5e-4}. The learning rate for $E$ is set to 0.05 initially and divided by 10 after every 40 epochs. The learning rate for other parts is $\frac{1}{10}$ of the learning rate for $E$. \algoref{phaseI} depicts the detailed training procedure of this phase.


\begin{algorithm}[!htb]
\caption{Phase I of the training procedure.}
\label{algo:phaseI}
\SetKwRepeat{Do}{do}{while}%
\KwIn{Training data $\{\bm x_i\}$ along with the identity label ${\bm c}_\mathcal{P}^i$ and the illumination label $c_\mathcal{I}^i$. Initialized parameters $\theta_\mathcal{P}$ and $\theta_\mathcal{I}$. Hyperparameters $\lambda_{1, 2, 3}$, $\xi$ and learning rate $\mu^t$. The number of iterations $t \gets 0$.}
\KwOut{Parameters  $\theta_{E}$, $\theta_\mathcal{P}$ and $\theta_\mathcal{I}$.}
\While{not converge}{
$t \gets t + 1$.\\
Compute the joint loss by $\mathcal{L}^t \!=\! \mathcal{L}_\personSym^t\!+\!\lambda_3\mathcal{L}_\lightSym^t$.\\
Compute the back-propagation error $\frac{\partial \mathcal{L}^t}{\partial \bm{x}_i^t}$ for each $i$ by $\frac{\partial \mathcal{L}^t}{\partial \bm{x}_i^t}\!=\!\frac{\partial \mathcal{L}_\personSym^t}{\partial \bm{x}_i^t}\!+\!\lambda_3\frac{\partial \mathcal{L}_\lightSym^t}{\partial \bm{x}_i^t}$.\\
Update the parameters $\theta_{E}$, $\theta_\mathcal{P}$ and $\theta_\mathcal{I}$
}
\end{algorithm}

\tabcolsep=4pt
\begin{figure}[t]
\centering 
\begin{tabular}{cc}
\includegraphics[width=0.45\columnwidth]{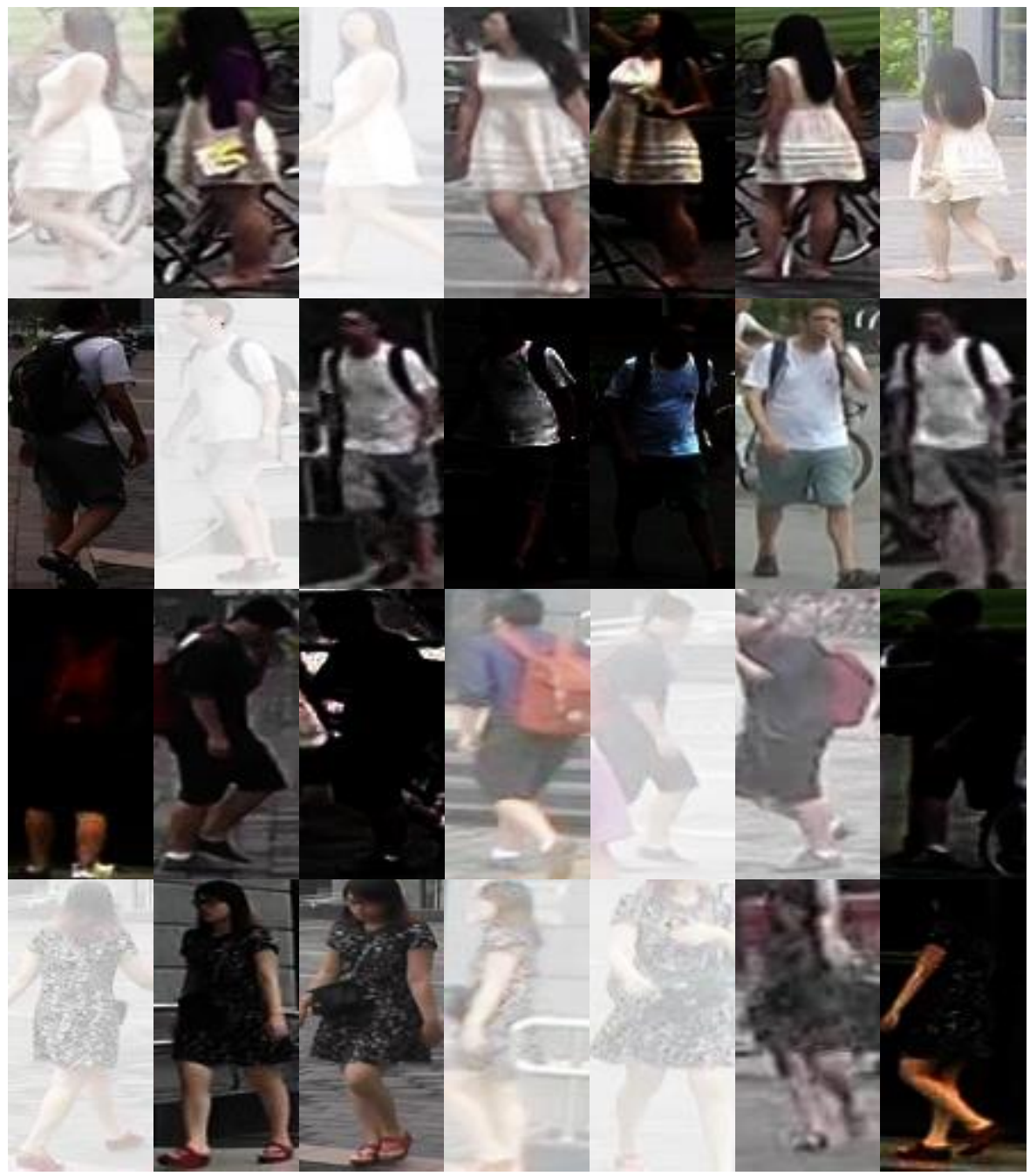} & %
\includegraphics[width=0.45\columnwidth]{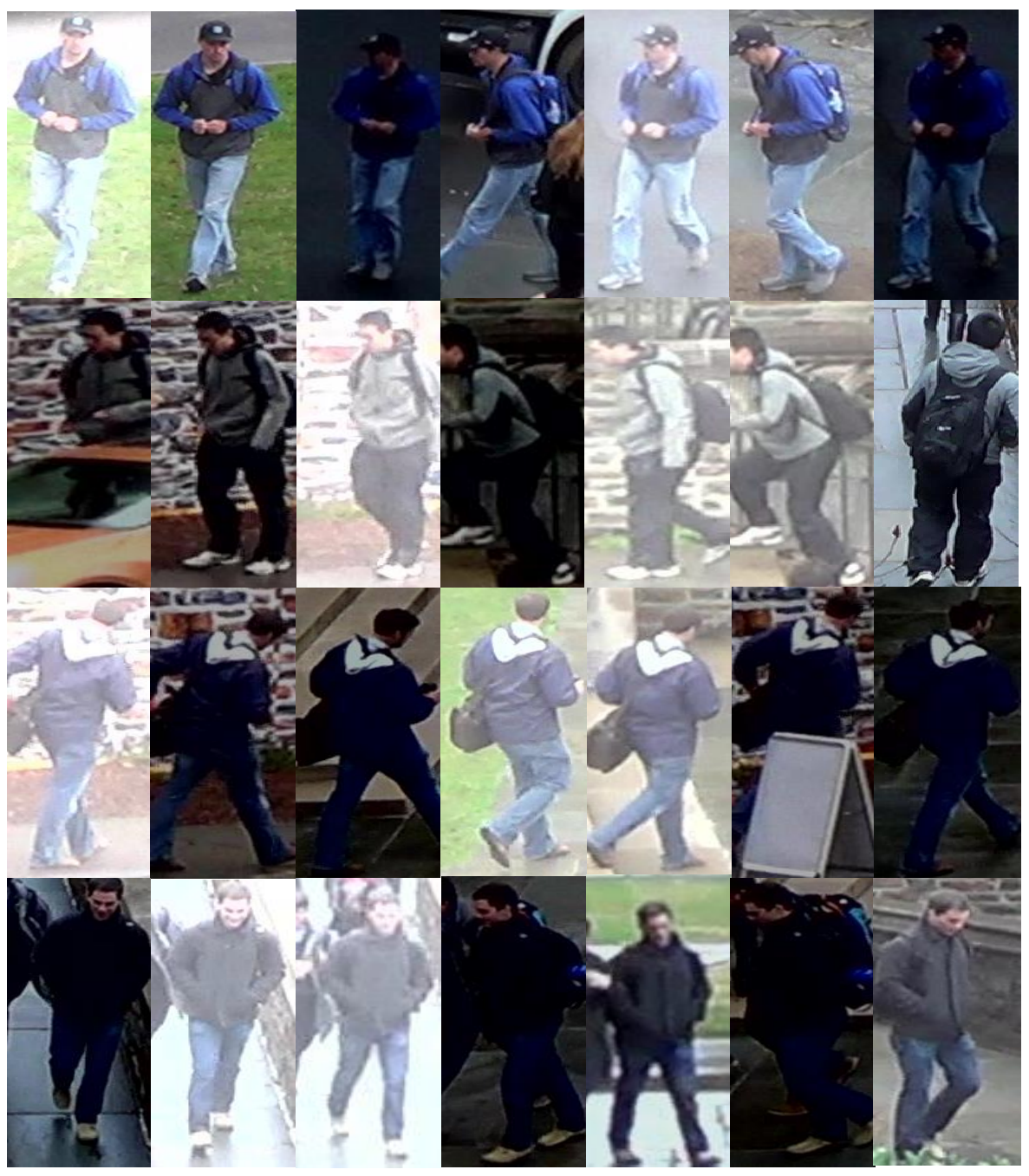} \\ %
\small{(a) Market-1501++} & \small{(b) DukeMTMC-reID++}
\end{tabular}
\caption{{\bf Sample images of the two simulated datasets.} Each row shows images of the same identity. For each identity, the images have different illuminations. (a) The Market-1501++ dataset. (b) The DukeMTMC-reID++ dataset.}
\label{fig:datasets}
\end{figure}

\heading{Phase II: Generator training.} With the disentangled feature learning modules $\{E, \mathcal{H}_{\personSym/\lightSym}, {\mathbf W}_{\personSym/\lightSym}, {\mathbf b}_{\personSym/\lightSym}\}$ fixed, we optimize the image generator $G$ with the Adam optimizer in phase II. The learning rate is set to \num{0.01} and reduced to 0.1 of its previous value every 40 epochs. The training batchsize is 64. Our task is to reconstruct the input images, \ie, minimize the \emph{MSE} loss. The objective function is 
\begin{equation}
\label{eq:gan_obj}
\argmin_{\theta_G} \mathcal{L}_\mathcal{G}\,,
\end{equation}
where $\theta_G$ represents parameters of the generator $G$.

\heading{Phase III: Joint training.} Finally, we jointly train the entire network in an end-to-end manner and the overall objective function is expressed as
\begin{equation}
\label{eq:overall_obj}
\argmin_{\theta_E, \theta_\personSym, \theta_\lightSym, \theta_G} \mathcal{L}_\personSym + \lambda_3\mathcal{L}_\lightSym + \lambda_4\mathcal{L}_\mathcal{G}\,.
\end{equation}
We use Adam optimizer for optimizing the overall objective function. The initial learning rate for the feature disentanglement part and the generator part is set to \num{1e-4} and  \num{1e-3} respectively. It decreases to 0.1 of its previous value every 50 epochs. The hyperparameters are $\lambda_3 \!=\! 1$ and $\lambda_4 \!=\! 2$. The generation process gifts the network more ability to disentangle the illumination feature. Hence, a joint learning manner will better balance the network's abilities of re-identification and disentanglement. 

\section{Experiments}
\label{sec:exp}

\subsection{Experimental Settings}
\label{subsec:dataset}
\textbf{Datasets.} There are two widely used datasets under normal illumination, \ie, Market-1501~\cite{zheng2015scalable} and DukeMTMC-reID~\cite{zheng2017unlabeled}. For Market-1501, it consists of 12,936 images of 751 identities for training and 19,281 images of 750 identities in the gallery set for testing. For DukeMTMC-reID, it contains 16,522 training images with 702 identities, 2,228 query images of the other 702 identities and 17,661 gallery images. 

Based on these two datasets, we constructed two simulated illumination-adaptive datasets. Considering that a slight illumination variation does not change the representation too much, and also that multiple scales of illuminations are required to simulate a wide range of illumination variation, we selected nine scales of illuminations. We adapt each image to a random one of nine illuminations. We apply a random gamma adjustment to each channel of the common images to produce the illumination-adaptive images, which is similar to~\cite{lore2017llnet}. The recorded value of an image captured by a camera is usually nonlinearly mapped from its corresponding scene radiance, and the nonlinearity often can be well approximated by a power function. The variance in the real-world illumination is then nonlinearly related to the image intensity. Therefore, we applied the nonlinear gamma transform to the decomposed illumination for simulating images under different illumination conditions. To make the illumination change reasonable, we further add Poisson noise with peak value = 10 to illumination changed images. Finally, we constructed the simulated illumination-adaptive datasets and named them as Market-1501++ and DukeMTMC-reID++. \figref{datasets} gives examples of these two simulated datasets. We consider that the gamma transform is insufficient to model non-global illumination variation. However, although local illumination variation could affect face identification significantly, person ReID relies mostly on global information and is consequently less sensitive to local illumination variation. Thus, we only consider global illumination variation.

\tabcolsep=10pt
\begin{table*}[!htb]
\renewcommand{\arraystretch}{1.15}
\caption{Comparison with the state-of-the-art methods on the Market-1501++ and DukeMTMC-reID++ datasets. CMC-1, CMC-5, CMC-10~(\%) and mAP~(\%) are reported.}
\label{tab:comp_sota}
\centering
\resizebox{\textwidth}{!}{
\begin{tabular}{l| c c c c | c c c c }\toprule
\multirow{2}{*}[0em]{\bf{Method}} & \multicolumn{4}{c|}{\bf{Market-1501++}} & \multicolumn{4}{c}{\bf{DukeMTMC-reID++}} \\
              & CMC-1 & CMC-5 & CMC-10 & mAP & CMC-1 & CMC-5 & CMC-10 & mAP\\\midrule
DenseNet121~\cite{huang2017densely}  &0.74  &2.29  &3.53  &0.73   &1.21 &2.74 &4.13 &0.80\\
DenseNet121 w/ Train              &70.60 &85.36 &89.66 &49.79  &64.45&77.82&82.45&45.12\\
PCB~\cite{sun2018beyond}          &0.56  &1.69  &2.91  &0.54   &0.72 &2.15 &3.23 &0.49\\
PCB  w/ Train                  &72.55 &85.22 &90.08 &53.11  &65.98&77.93&82.21&45.15\\
ResNet50~\cite{he2016deep}           &0.42  &1.16  &2.05  &0.39   &0.54 &1.97 &3.14 &0.50\\
ResNet50  w/ Train (Baseline)     &66.18 &81.97 &87.02 &47.71  &62.07&75.54&88.08&42.63\\
\midrule
IID                               &{\bf 73.37} &{\bf 86.55} &{\bf 91.01} &{\bf 56.22}                                                     &{\bf 68.11} &{\bf 79.75} &{\bf 91.27} &{\bf 49.20}\\
Improvement over baseline          &\textcolor{blue}{7.19$\uparrow$} &\textcolor{blue}{4.58$\uparrow$} &\textcolor{blue}{3.99$\uparrow$} &\textcolor{blue}{8.51$\uparrow$}   &\textcolor{blue}{6.04$\uparrow$} &\textcolor{blue}{4.21$\uparrow$} &\textcolor{blue}{3.19$\uparrow$} &\textcolor{blue}{6.57$\uparrow$}\\
\bottomrule
\end{tabular}}
\end{table*}

\tabcolsep=10pt
\begin{table*}[!htb] 
\caption{Ablation study on the Market1501++ dataset. CMC-1 (\%) and mAP (\%) are reported.}
\centering
\label{tab:ablation}
\resizebox{0.7\textwidth}{!}{
\begin{tabular}{l| c c c c | c c }\toprule
\multirow{2}{*}[0em]{\bf{Method}} & \multicolumn{4}{c|}{\bf{Components}} & \multicolumn{2}{c}{\bf{Market-1501++}} \\
              & $\mathcal{L}_{\personSym}^T$ & $\mathcal{L}_{\personSym}^S$  & $\mathcal{L}_{\mathcal{I}}$ & $\mathcal{L}_\mathcal{G}$  & CMC-1 & mAP \\\midrule
Baseline      & \cmark & \cmark & \xmark & \xmark  &66.18 &47.71 \\
IID (no $G$)  & \cmark & \cmark & \cmark & \xmark  &71.54 &55.17 \\
IID (no triplet for id)   & \xmark & \cmark & \cmark & \cmark  &64.14 &45.87 \\
IID (no softmax for id)   & \cmark & \xmark & \cmark & \cmark  &65.21 &46.53 \\
IID (no illum.)   & \cmark & \cmark & \xmark & \cmark  &70.79 &54.57 \\
IID           & \cmark & \cmark & \cmark & \cmark  &73.37 &56.22 \\
\bottomrule
\end{tabular}}
\end{table*}

\heading{Evaluation metrics.}
To indicate the performance, the standard Cumulative Matching Characteristics (CMC) values and mean Average Precision (mAP) are adopted~\cite{zheng2015scalable}, since one person has multiple ground truths in the gallery set.

\begin{figure}[!tb]
\centering
    \includegraphics[width=0.7\columnwidth]{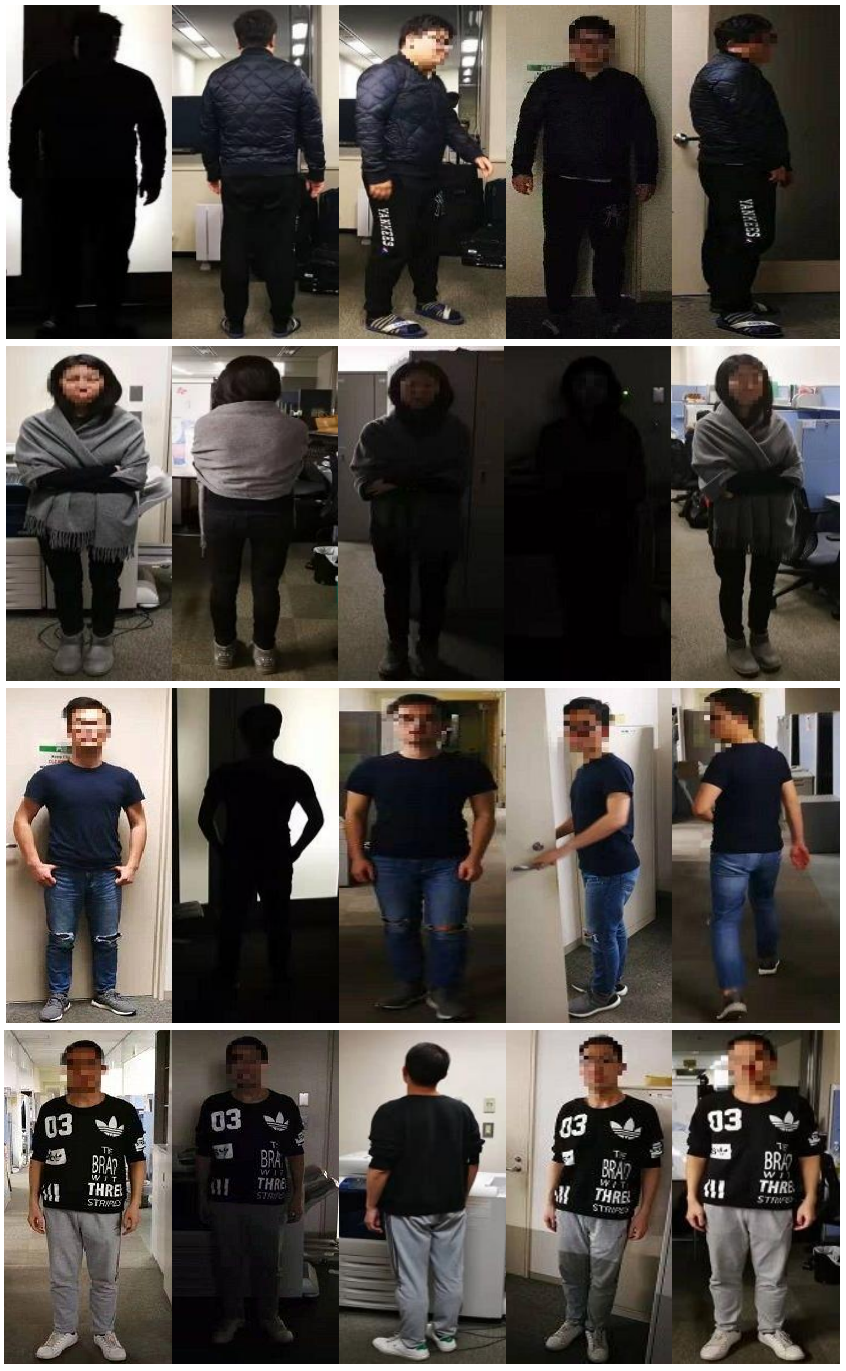}
  \caption{{\bf Sample images of the collected real-world images.} Each row shows images of one identity.}
  \label{fig:samples}
\end{figure}

\heading{Real-world images.} To prove the effectiveness of the proposed method on reducing the effect of illuminations, we have also collected some real-world images with different illuminations. Note that we collected real-world images indoors. Because of the controllable indoor environment, we can collect diverse real-world images easily. We will collect more real-world images outdoors (at morning, noon and nightfall) in the future. Through calculating the distances, we show the ability of the proposed IID network. Some examples are shown in \figref{samples}. 

\begin{figure}[!tb]
\centering
\begin{tabular}{cc}
    \includegraphics[width=0.49\columnwidth]{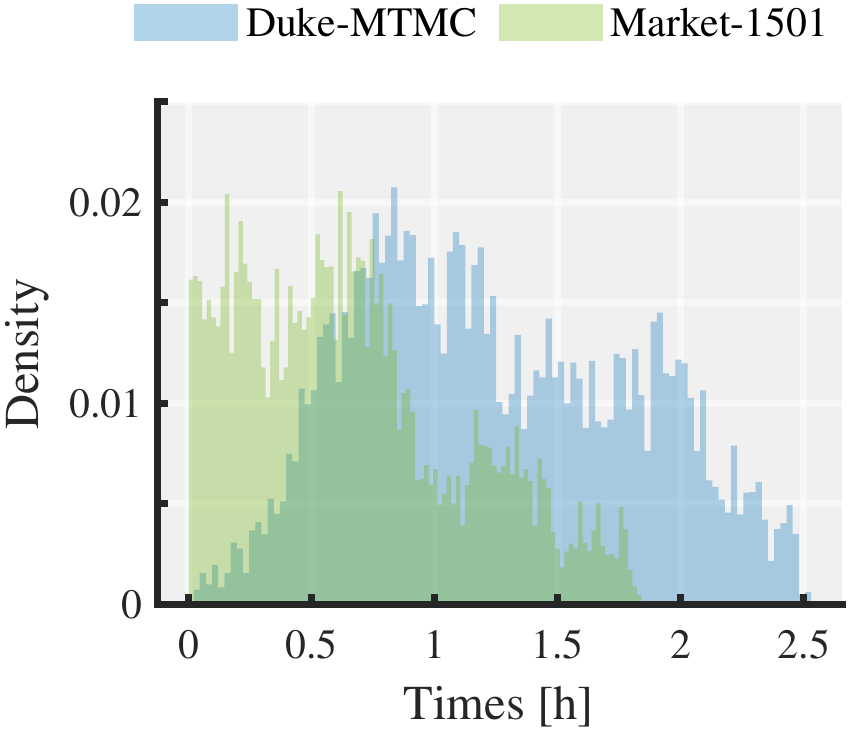} &
    \includegraphics[width=0.49\columnwidth]{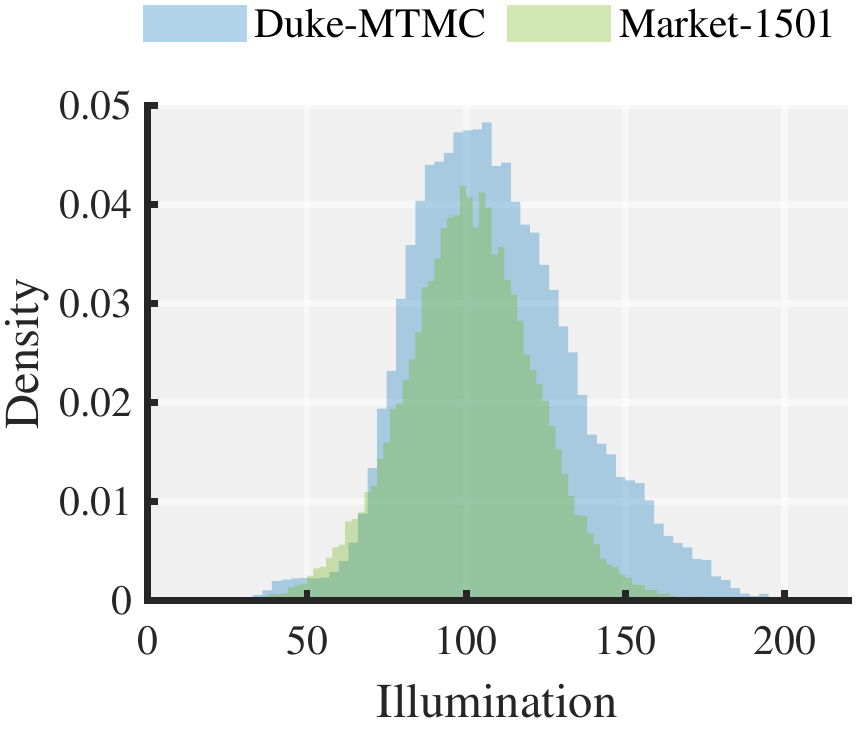}\\
    \multicolumn{2}{c}{\includegraphics[width=\columnwidth]{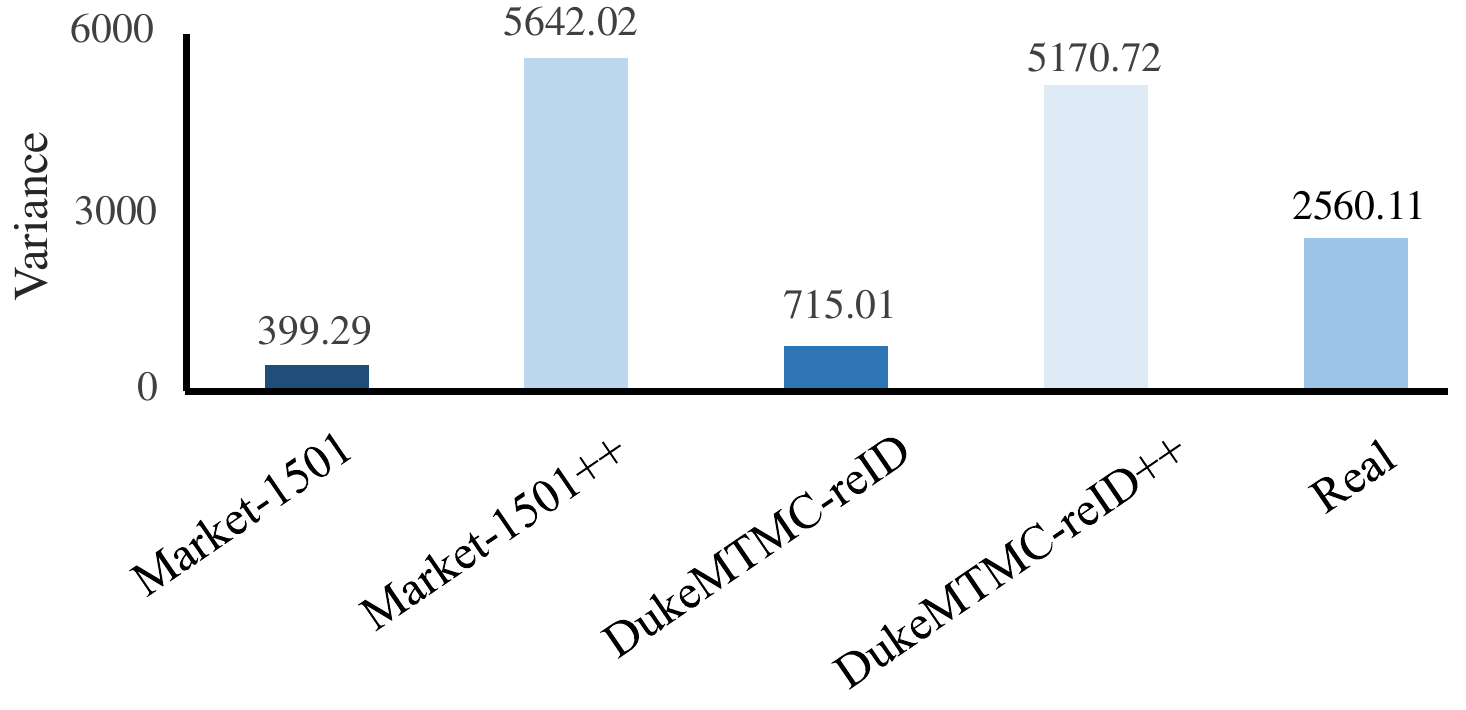}}
\end{tabular}  
  \caption{{\bf Illumination analysis of related datasets.} We use the average value of the pixel luminance of each image to represent its illumination. The top two figures show the illumination distributions of Market-1501 and DukeMTMC-reID datasets. We can find that images were captured around one and two hours respectively, and their illuminations do not change too much from the mean value 100. The bottom figure shows the variance of the image illumination on related datasets. We can find that the illumination variances of the simulated datasets (Market-1501++ and DukeMTMC-reID++) and the real dataset are much larger than that of the existing ReID datasets (Market-1501 and DukeMTMC-reID).}
  \label{fig:mlv}
\end{figure}

For analyzing illumination conditions of related datasets, we use the average luminance value of each image to represent its illumination and investigate the illumination distributions of the Market-1501 and DukeMTMC-reID datasets. We also made a comparison of the variance of the image illumination on related datasets. \figref{mlv} shows the results. We can find that the illumination variances of the simulated datasets and the real dataset are much more significant than those of the existing ReID datasets. Hence, existing ReID datasets were captured in a short period of time, and the images have relatively uniform illumination conditions. We introduce a person re-identification task that has more significant illumination variations and requires better illumination adaption.

\tabcolsep=1pt

\subsection{Comparison with State-of-The-Arts}
In this subsection, we make comparisons with the state-of-the-art methods. We exploit the Market-1501++ and the DukeMTMC-reID++ datasets to evaluate the methods. As IA-ReID is new, there are barely methods for comparisons. We selected DenseNet121~\cite{huang2017densely}, PCB~\cite{sun2018beyond} and ResNet50~\cite{he2016deep} as the comparison methods. DenseNet121 and ResNet50 are two popular baseline networks in ReID. PCB is one of the state-of-the-art ReID methods. \tabref{comp_sota} list their results. Note that the notation with a `w/ Train' suffix means that the indicated model has been trained on the illumination-adaptive datasets before testing. From the table, we can find that the results of DenseNet121, PCB and ResNet50 drop dramatically when dealing with the IA-ReID datasets. However, all of these three deep networks can receive very high promotions when training on the illumination-adaptive data, which means that the deep learning network can somehow deal with part of the illumination-adaptive issue if given proper training data. 

We took ResNet50 as our baseline. From \tabref{comp_sota}, we can find that comparing with the baseline (serving as our backbone network), our method improves the performance on both the simulated Market-1501++ and DukeMTMC-reID++ datasets. Thus, The improvement against the baseline better shows the effectiveness of the proposed method as they share the same backbone architecture. Although we do not report improvements against other methods, from \tabref{comp_sota}, it is clear that our method outperforms all methods. Note that our method uses a very basic baseline (ResNet50), and makes a considerable improvement (8.51\% mAP on the Market-1501++ dataset and 6.57\% mAP on the DukeMTMC-reID++ dataset). However, compared with PCB (designed to address the misalignment challenge in general ReID tasks), our method does not pay attention to the misalignment problem, which still exists in the ReID task, making our improvement less remarkable. We consider that if we design a new module for the misalignment problem as PCB does, our method will outperform PCB with a much larger gain.

\tabcolsep=12pt
\begin{table}[!htb] 
\caption{Comparison with the baseline on the Market-1501 and DukeMTMC-reID datasets. CMC-1 (\%) and mAP (\%) are reported.}
\centering
\label{tab:reid-data}
\resizebox{\columnwidth}{!}{
\begin{tabular}{l| c c | c c }\toprule
\multirow{2}{*}[0em]{\bf{Method}} & \multicolumn{2}{c|}{\bf{Market-1501}} & \multicolumn{2}{c}{\bf{DukeMTMC-reID}} \\
              & CMC-1 & mAP  & CMC-1 & mAP \\\midrule
Baseline      &88.84 &71.49  &79.71 &61.77 \\
IID           &88.45 &71.46  &78.10 &60.56 \\\bottomrule
\end{tabular}}
\end{table}

\subsection{Ablation Study}
Our method consists of four kinds of losses. The triplet loss $\mathcal{L}_{\personSym}^T$ and the cross-entropy loss $\mathcal{L}_{\personSym}^S$ are responsible for extracting robust person features. The regression loss $\mathcal{L}_{\mathcal{I}}$ is responsible for predicting proper illumination features. The \emph{MSE} loss $\mathcal{L}_\mathcal{G}$ is used for image reconstruction. Removing $\mathcal{L}_{\personSym}^T$ or $\mathcal{L}_{\personSym}^S$ will give less constraint to the person feature, and thus degrade the re-identification performance. $\mathcal{L}_{\mathcal{I}}$ and $\mathcal{L}_\mathcal{G}$ influence the effectiveness of illumination disentanglement. Removing either of them will degrade the re-identification performance indirectly.

Here, we take the Market-1501++ dataset for the ablation study. When removing both of $\mathcal{L}_\mathcal{I}$ and $\mathcal{L}_\mathcal{G}$ (the baseline), the performance drops to $66.18\%$ CMC-1 and $47.71\%$ mAP, as it can only rely on the person feature without separating the illumination feature. When removing $\mathcal{L}_\mathcal{G}$, the performance does not drop so much as the baseline. The loss $\mathcal{L}_{\mathcal{I}}$ is useful for separating the illumination information and promotes the re-identification result. When removing $\mathcal{L}_{\personSym}^T$ or $\mathcal{L}_{\personSym}^S$, the performance drops dramatically to be even worse than the baseline. So the loss for re-identification is essential for the IA-ReID task. When removing $\mathcal{L}_{\mathcal{I}}$, the performance does not drop so much as the baseline. So even without the loss of illumination regression, the generation process can also benefit the illumination disentanglement.

\subsection{Experiments on General ReID Datasets}
Although we design a new network for the IA-ReID task, we do not expect the proposed IID network performing poorly on the general ReID dataset. As our network is proposed based on the baseline network Resnet50, we make a comparison with the baseline. The results are listed in \tabref{reid-data}. We can find that the results do not change too much. Although the IID network is specially designed for the illumination-adaptive condition, it is still suitable for the general ReID task. 

\begin{figure}[!tb]
\centering
    \includegraphics[width=\columnwidth]{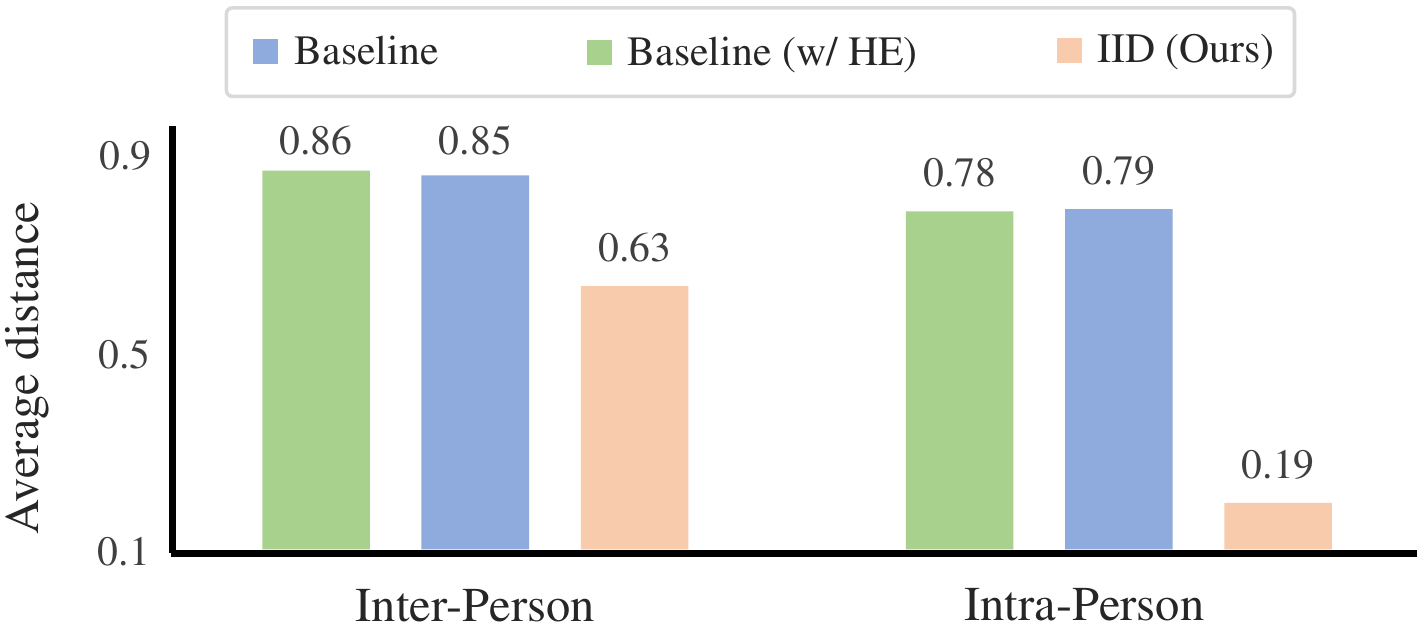}
  \caption{{\bf The average values of intra-person distances and inter-person distance for real-world images.} The experiments are respectively conducted by the baseline method, the baseline method fed with image after histogram equalization (w/ HE) and our IID method.}
  \label{fig:real}
\end{figure}

\subsection{Experiments on Illumination Prediction}
As we know, the disentangled illumination feature can be used to predict the illumination scale of each image. We test its prediction accuracy on the Market-1501++ and DukeMTMC-reID++ datasets. The accuracy values are very high, which are 98.74\% and 98.53\% respectively for the Market-1501++ and DukeMTMC-reID++ datasets. It means that the disentangled illumination feature can well represent the illumination scale.

\subsection{Experiments on Local Illumination Change}
The synthetic datasets above are generated by the global illumination change. In this subsection, we generate datasets with local illumination changes and investigate the effectiveness of the proposed method. 

\begin{figure*}[!h]
    \centering
    \tabcolsep=3pt
    \begin{tabular}{cccc}
        \includegraphics[width=0.23\textwidth]{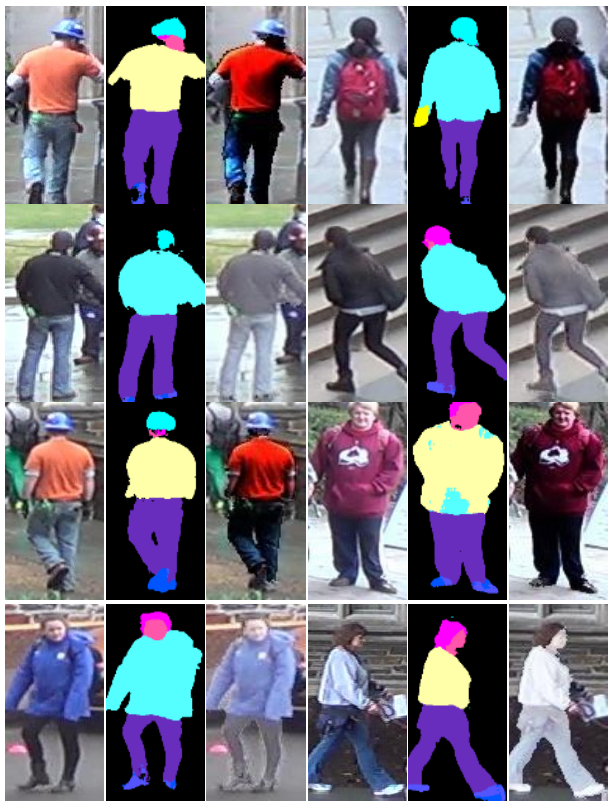} &
        \includegraphics[width=0.228\textwidth]{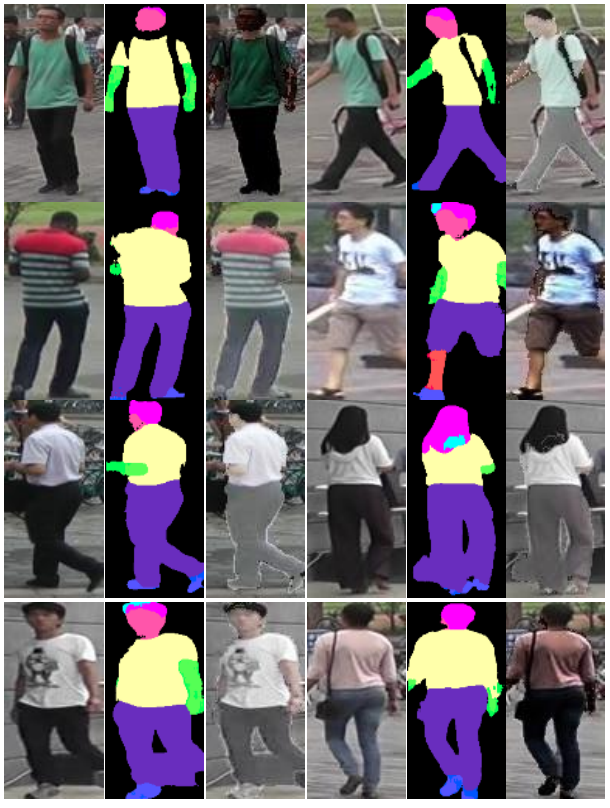}&
        \includegraphics[width=0.234\textwidth]{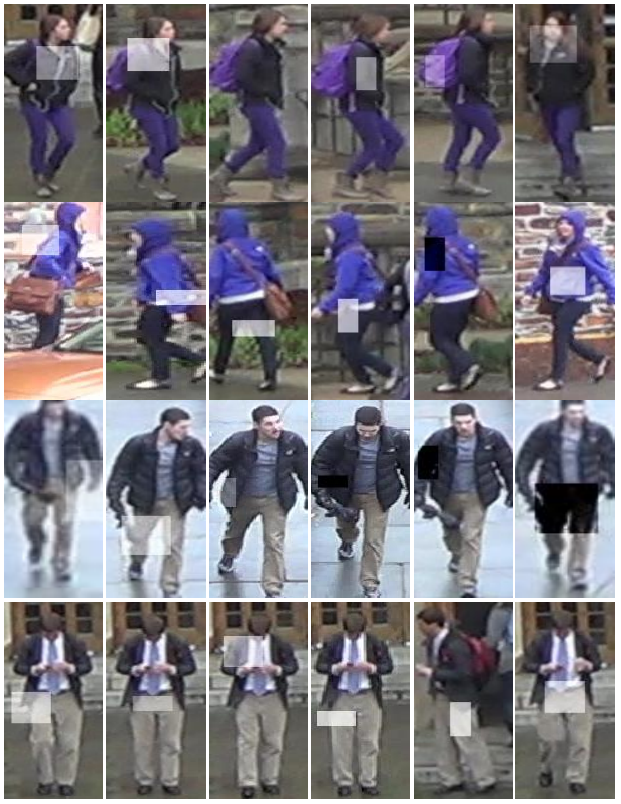} &
        \includegraphics[width=0.23\textwidth]{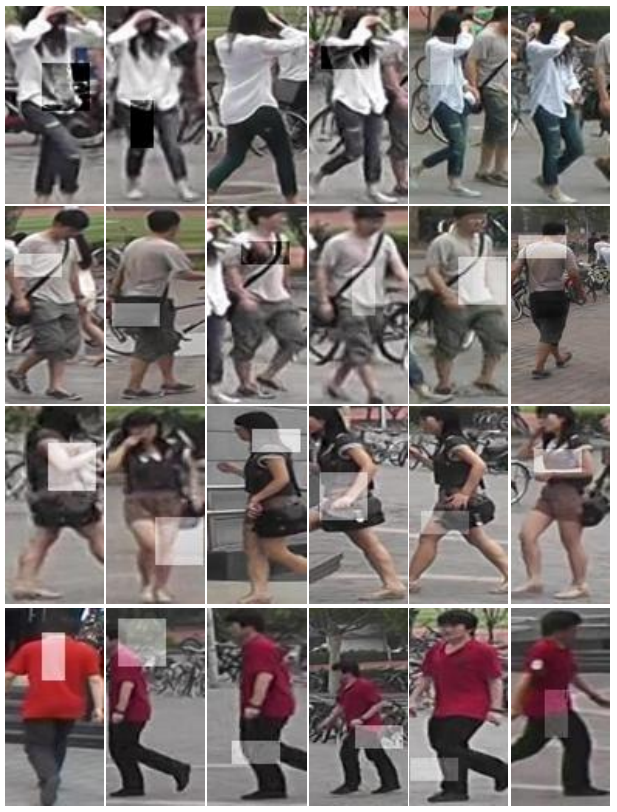} \\
        \small{(a) Foreground Duke} & \small{(b) Foreground Market} & \small{(c) Patch Duke} & \small{(d) Patch Market}\\
    \end{tabular}
    \caption{Examples of synthetic datasets with local illumination changes. (a) and (b) show samples by the foreground-based illumination change on the Duke and Market datasets, respectively. The images are respectively the original image, the person segmentation result, and the corresponding generated image. (c) and (d) show samples by the patch-based illumination change on the Duke and Market datasets, respectively.}
    \label{fig:local_examples}
\end{figure*}

We have tried to synthesize local illumination changes in two ways. 1) We exploite a person segmentation method~\cite{luo2018macro} to obtain the foreground person of each image and then change its illumination by using gamma correction. We set seven different gamma values. We name this synthesis process as foreground-based illumination change. Some generated examples are shown in \figref{local_examples}(a) and \figref{local_examples}(b). 2) We randomly select a patch in each image and then change its illumination by using gamma correction. We use three different sizes of patches, \ie, $32\!\times\!32$, $32\!\times\!64$ and $64\!\times\!32$. We also set seven different gamma values. We name this synthesis process as patch-based illumination change. Some generated examples are shown in \figref{local_examples}(c) and \figref{local_examples}(d). We exploit these two kinds of synthesis processes to generate new Market and Duke datasets, respectively. 

\begin{figure*}[!h]
    \centering
    \tabcolsep=3pt
    \begin{tabular}{cccc}
        \includegraphics[width=0.23\textwidth]{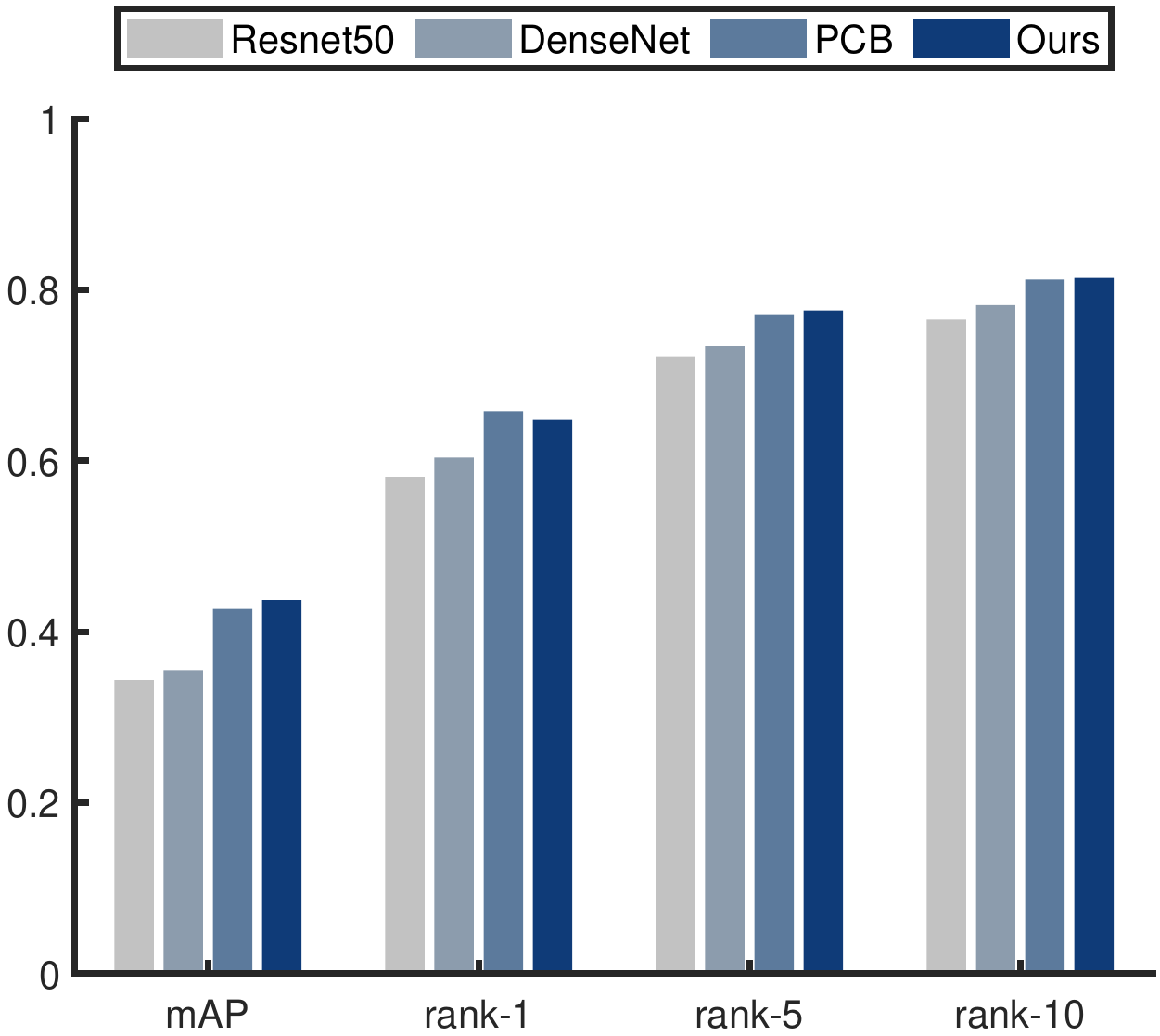} &
        \includegraphics[width=0.23\textwidth]{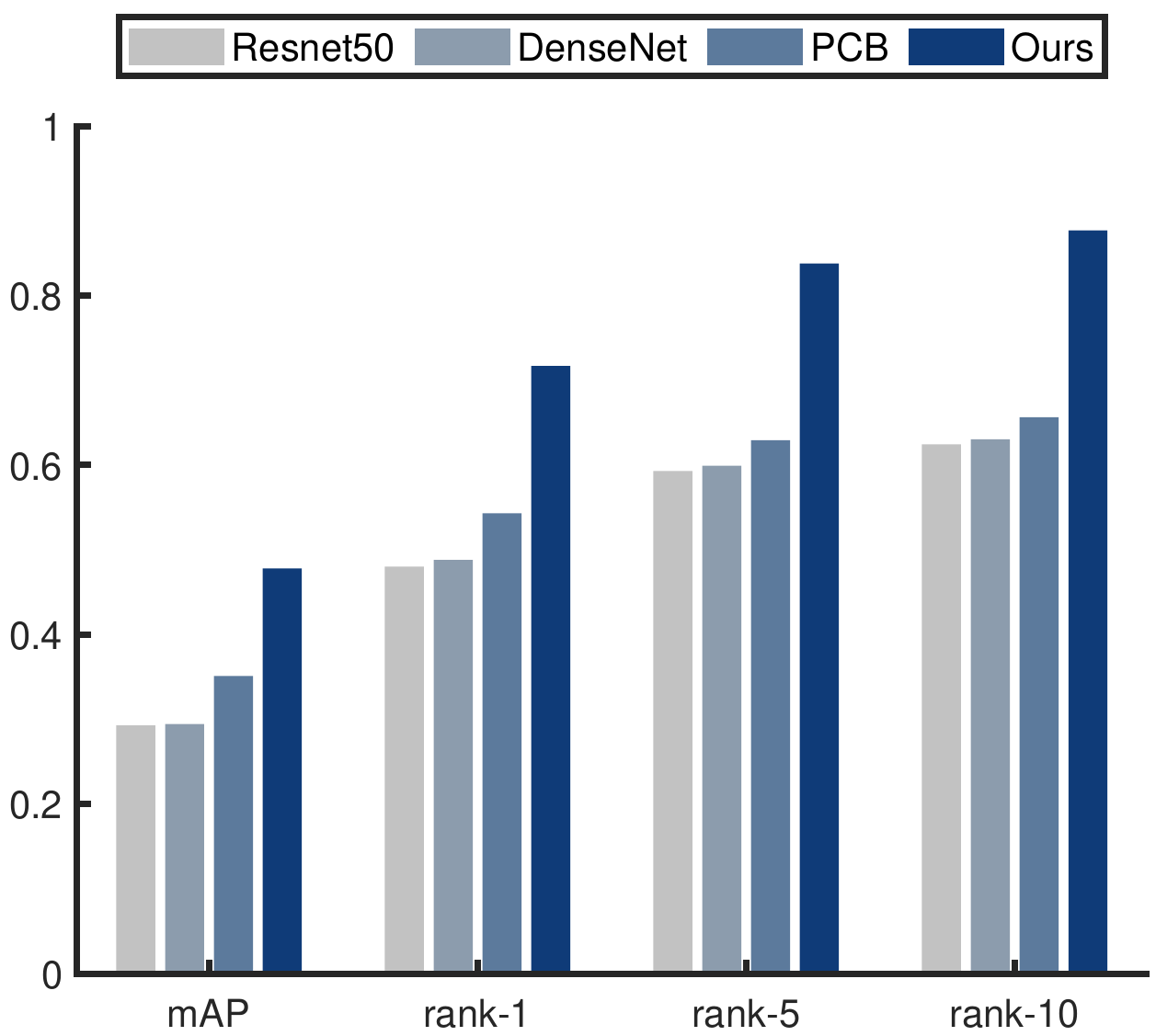}&
        \includegraphics[width=0.23\textwidth]{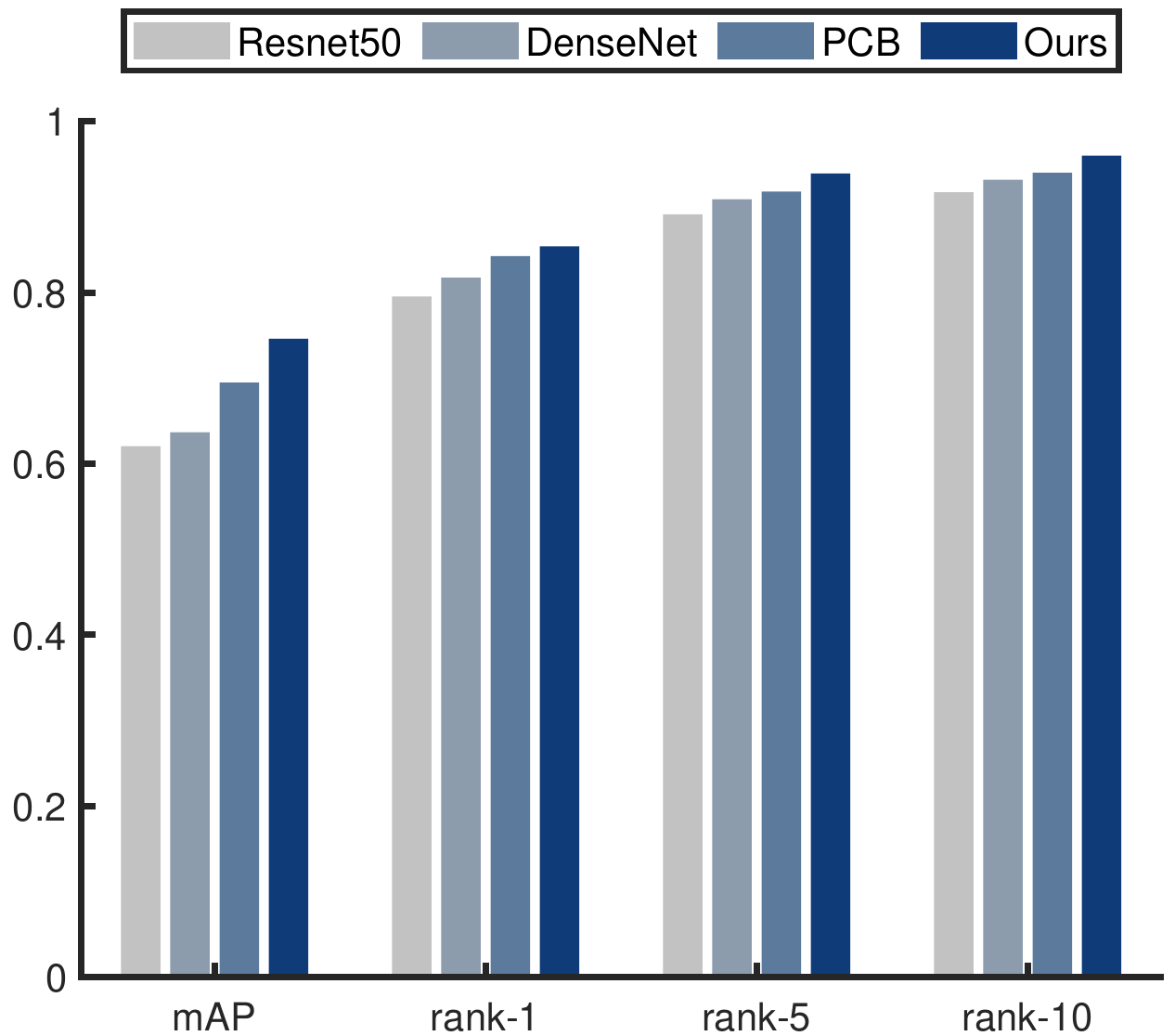} &
        \includegraphics[width=0.23\textwidth]{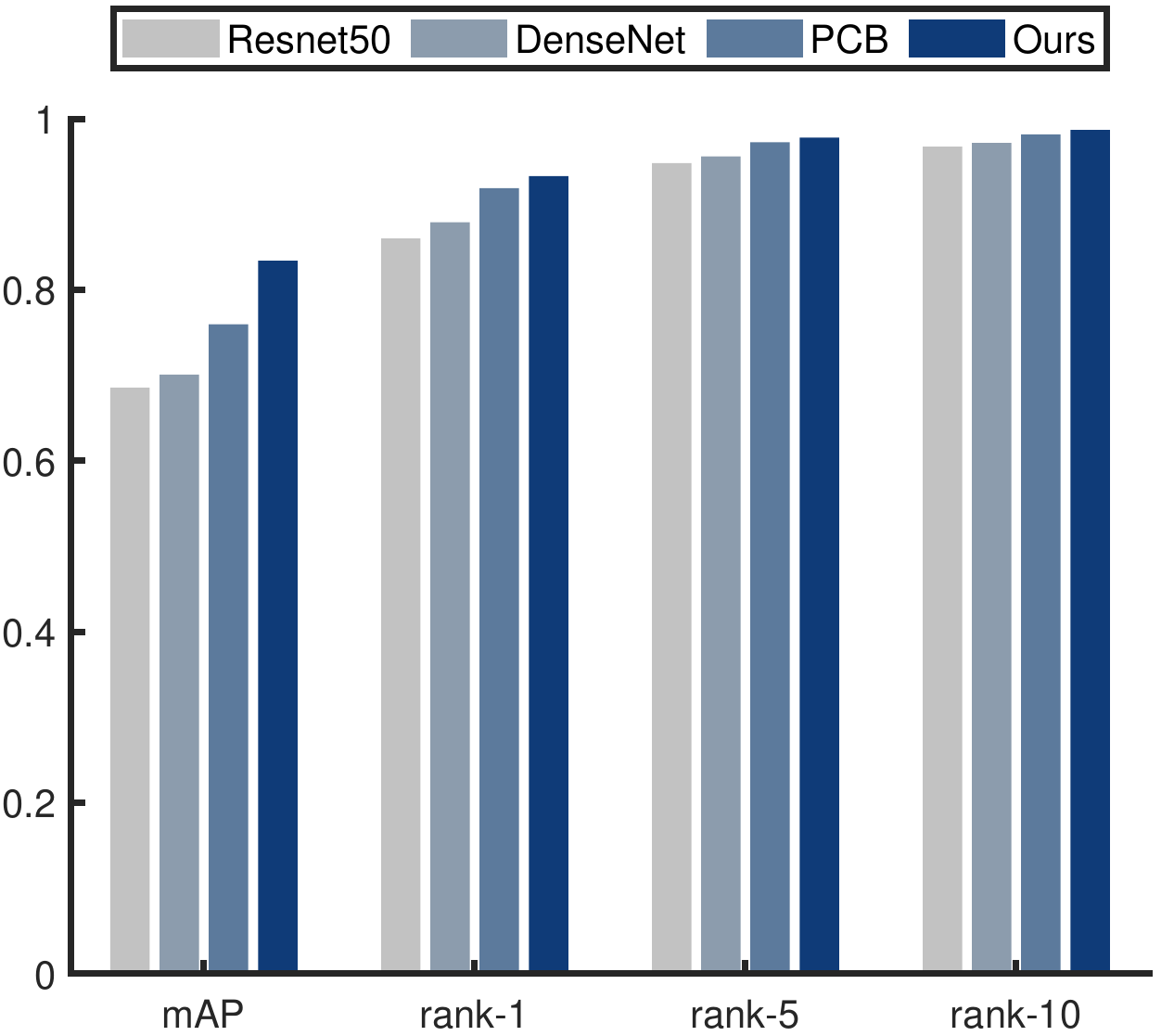} \\
        \small{(a) Foreground Duke} & \small{(b) Foreground Market} & \small{(c) Patch Duke} & \small{(d) Patch Market}\\
    \end{tabular}
    \caption{ReID results on four synthetic datasets with local illumination changes.}
    \label{fig:local_results}
\end{figure*}

Then, we conduct experiments on these four synthetic datasets with local illumination changes. We compare our methods with ResNet50, DenseNet, and PCB. \figref{local_results} shows the results, which demonstrates that the proposed method is more effective in the condition of local illumination changes.

\subsection{Experiments on the Real-world Images}
We recruited 15 volunteers, and for each volunteer, we collected ten images under different illumination conditions. Some examples are shown in \figref{samples}. We calculated the intra-person distances and inter-person distances respectively with the baseline model, the baseline model with histogram equalization, and the proposed IID network. \figref{real} shows the average intra-person and inter-person distances calculated by different methods. We can find that the proposed IID is more effective in reducing the intra-person distance, \ie, to alleviate the effect of the illumination change.

\begin{figure}[!h]
    \centering
        \includegraphics[width=0.6\columnwidth]{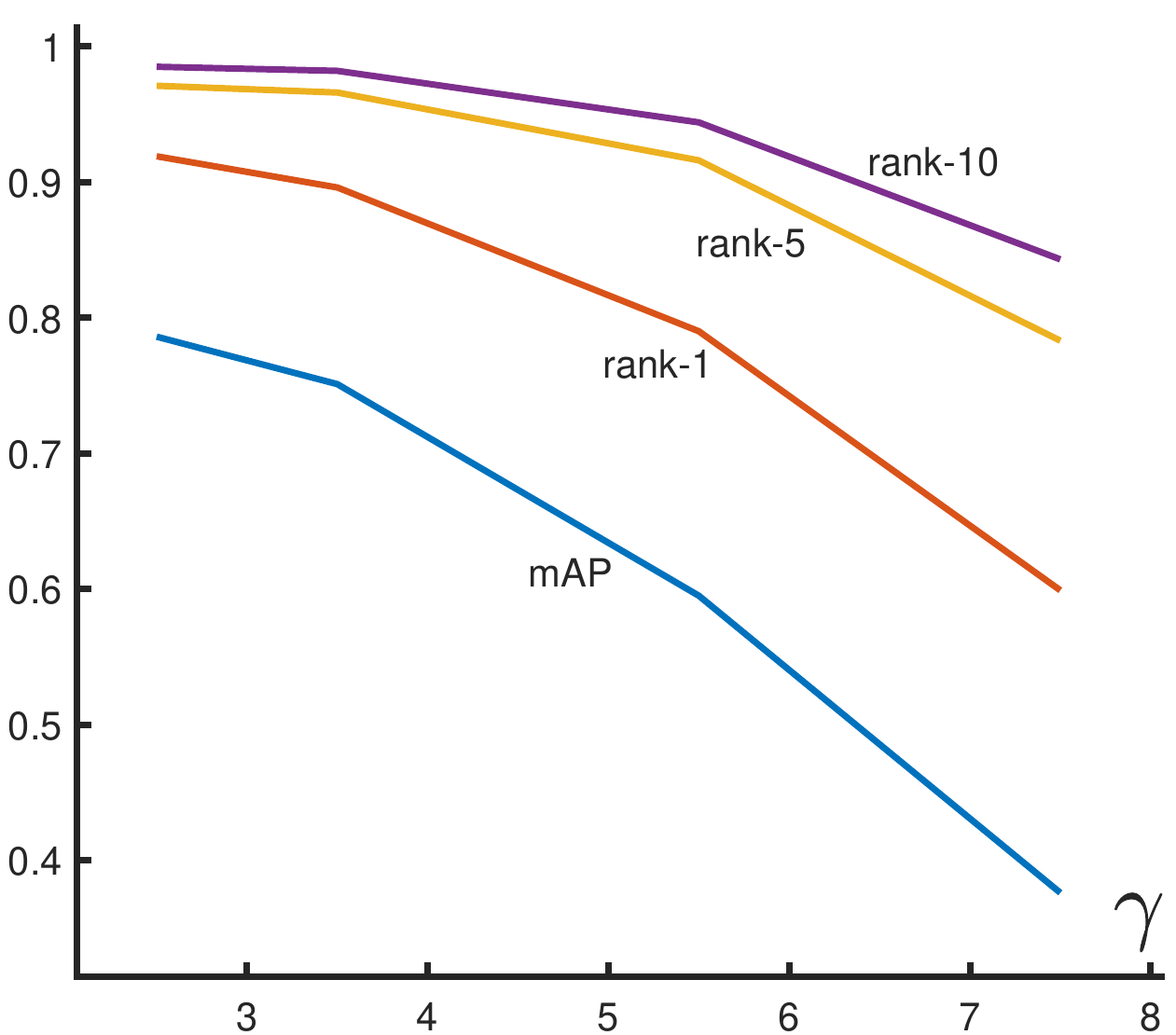}
    \caption{Results on low illumination conditions.}
    \label{fig:ill}
\end{figure}

\subsection{Experiments on Low Illumination Conditions}
In this subsection, we report experiments on low illumination images. We created four low-illumination testing sets based on Market-1501 by using four gamma corrections. Our model was trained on the Market-1501++ dataset (random illuminations) and tested on these four testing sets. \figref{ill} shows the results. We can find that the performances decrease as the illumination becomes lower. We consider that some useful information disappears and some noise is introduced in very low illumination conditions, and the proposed model can not extract enough useful information from the input image. However, in an appropriate low illumination range, our method can still work well.

\subsection{Experiments on Different Parameters}
To investigate the parameter $\lambda_3$ and $\lambda_4$ in the Equation (9), we conducted experiments on the Market-1501++ dataset with 1) fixed $\lambda_4$ and different $\lambda_3$ values, and 2) fixed $\lambda_3$ and different $\lambda_4$ values. \tabref{parameters} reports the results. We can find that when $\lambda_3=1$ and $\lambda_3=2$, our model achieves the best result. 

\begin{table}[h]
    \centering
    \tabcolsep=6pt
    \resizebox{0.4\columnwidth}{!}{
    \begin{tabular}{c|c|c}
    \toprule
          $\lambda_3$ & $\lambda_4$ & mAP(\%)\\\midrule
       0.1&  1.0 &  44.78\\
       0.5&  1.0 &  50.44\\
       1.0  &  1.0 & 54.31\\
       2.0  &  1.0 & 53.81 \\
       5.0  &  1.0 & 52.75 \\\midrule
       1.0  &  0.1& 53.19 \\
       1.0  &  0.5& 53.31 \\
       $\boldsymbol{1.0}$ &  $\boldsymbol{2.0}$& $\boldsymbol{56.22}$ \\
       1.0&  5.0& 48.47 \\
      \bottomrule
    \end{tabular}
    }
        \begin{tabular}{c}
         \includegraphics[width=0.5\columnwidth]{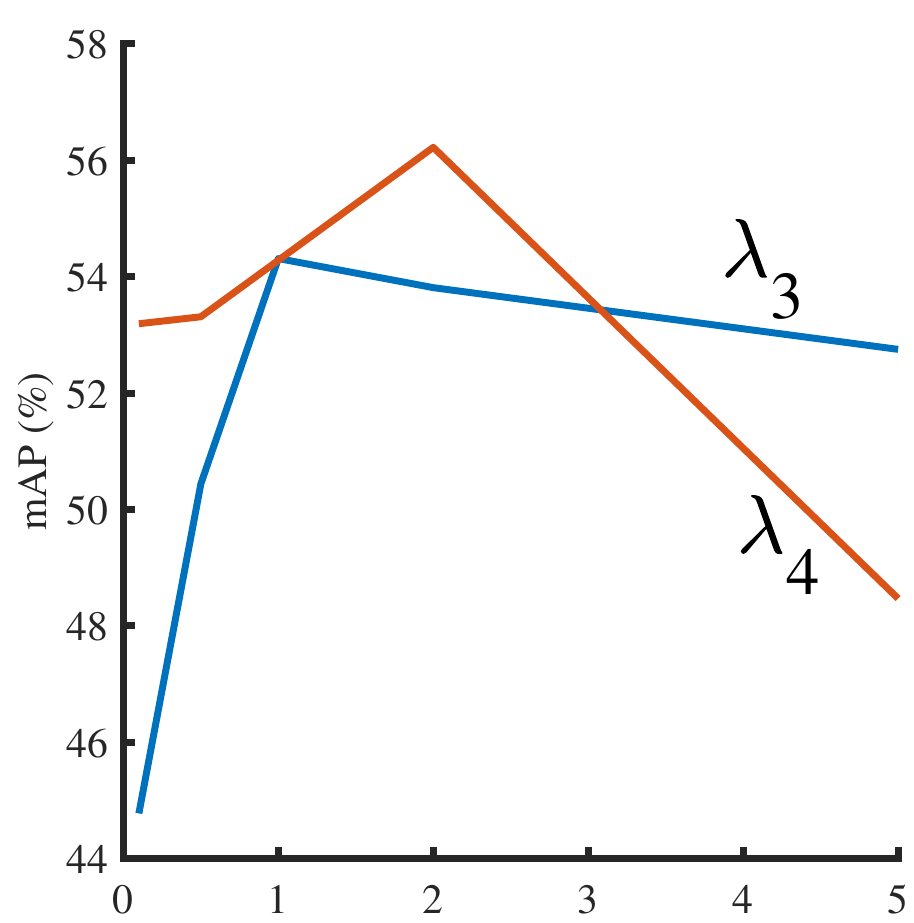}
    \end{tabular}
    \caption{Results on different parameters $\lambda_3$ and $\lambda_4$.}
    \label{tab:parameters}
\end{table}

\begin{figure*}[!h]
    \centering
    \tabcolsep=3pt
    \begin{tabular}{cccc}
            \includegraphics[width=0.23\textwidth]{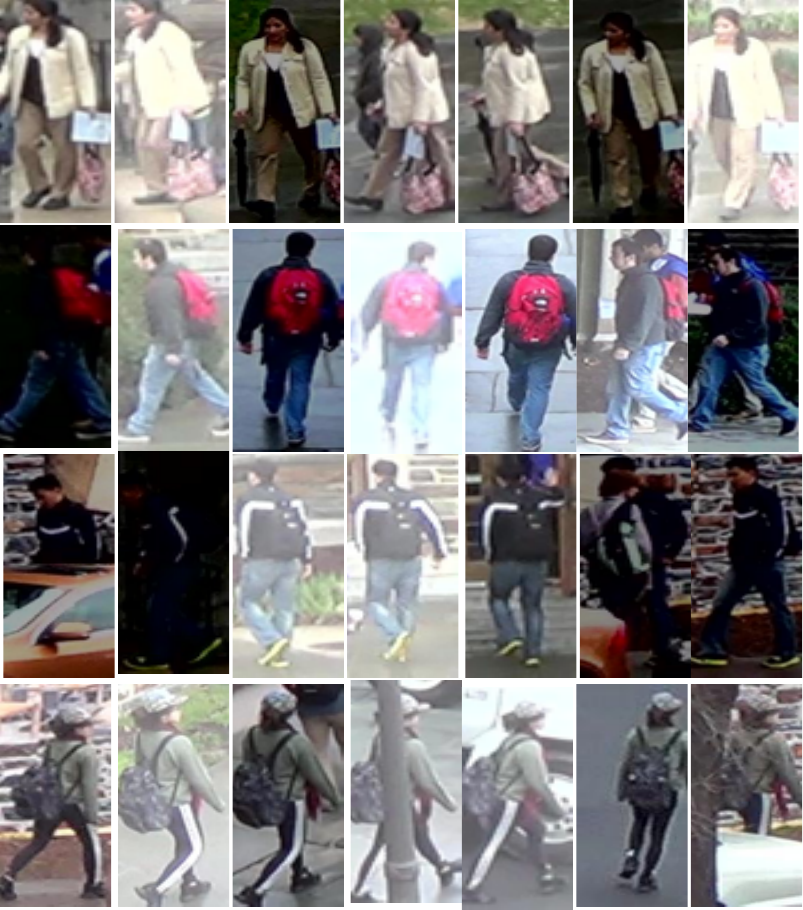} &
        \includegraphics[width=0.23\textwidth]{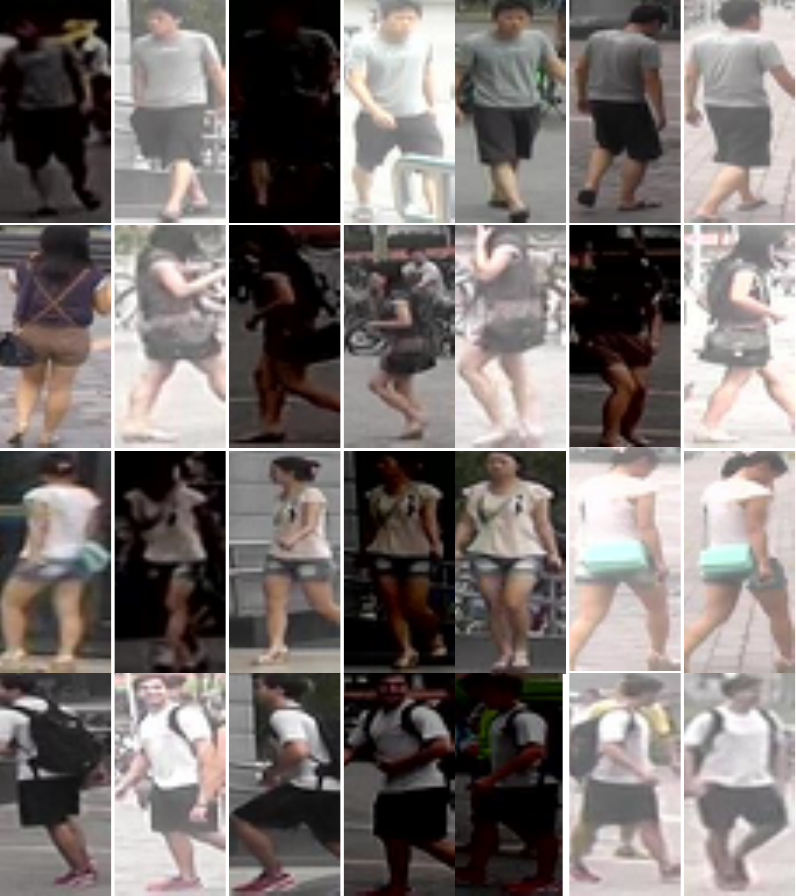} &
        \includegraphics[width=0.23\textwidth]{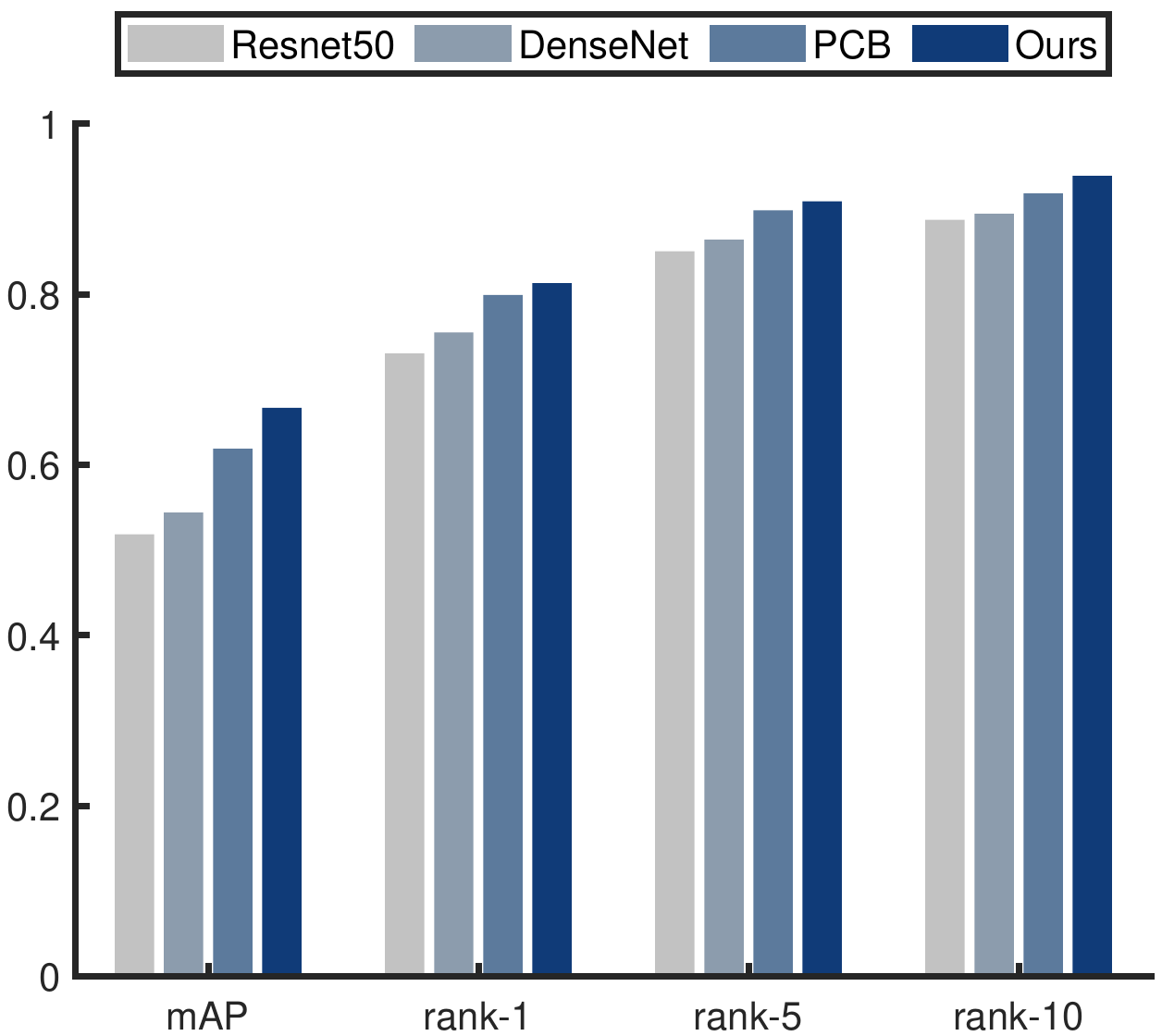} &
        \includegraphics[width=0.23\textwidth]{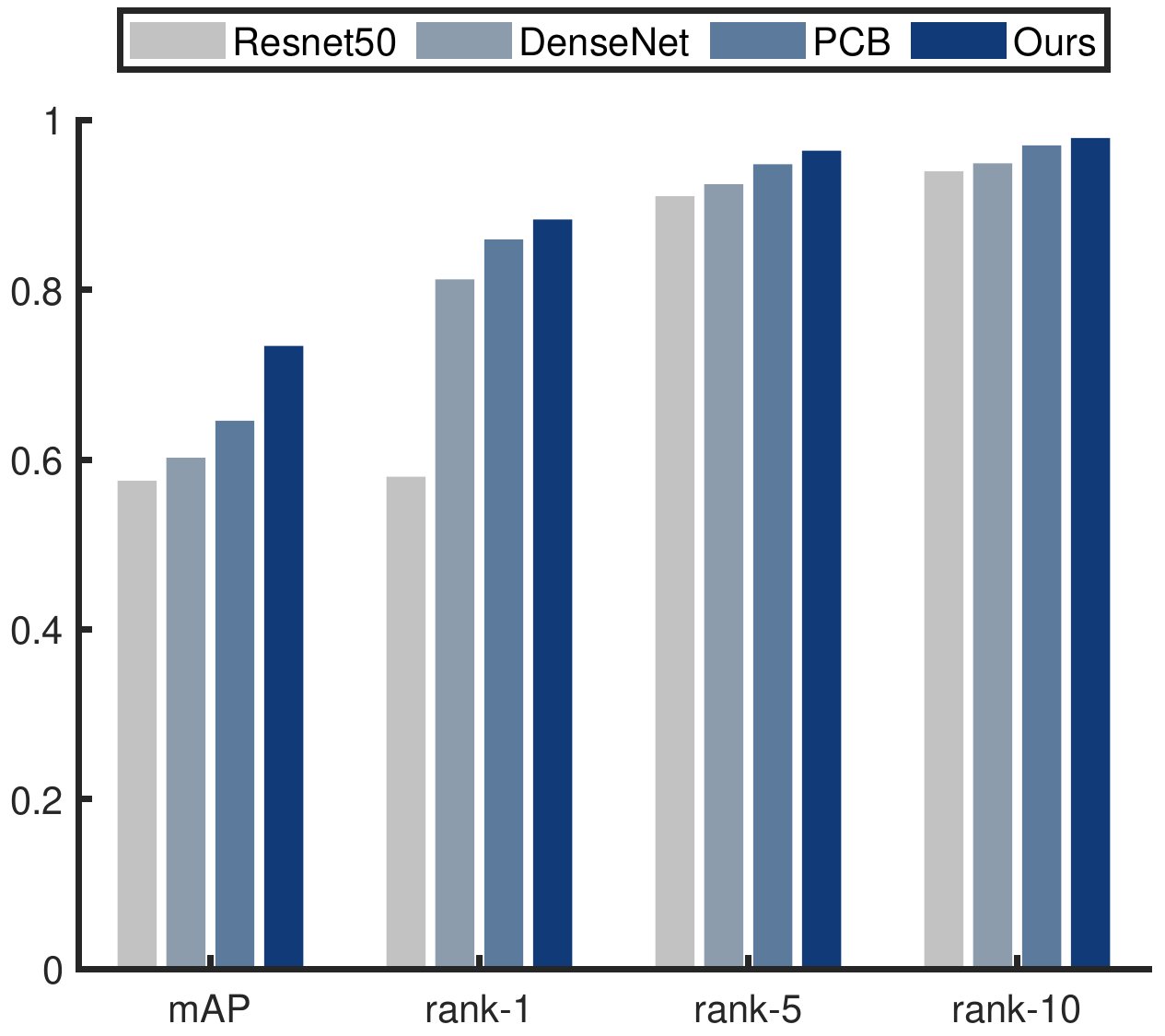}\\
        \small{(a) Duke Samples} & \small{(b) Market Samples} & \small{(c) Duke Results} & \small{(d) Market Results}\\
    \end{tabular}
    \caption{Examples and results of synthetic datasets with $Y$ channel illumination changes. (a) and (b) show samples by the $YUV$ illumination change on the Duke and Market datasets, respectively. (c) and (d) show the ReID results.}
    \label{fig:local_examples2}
\end{figure*}

\subsection{Experiments on Illumination-adaptive Datasets with $Y$ Channel Changes}
We exploit a non-parametric simulation method to generate images with different illuminations. We transfer the format of the image from $RGB$ to $YUV$. The $Y$ channel often reflects the illumination condition, and we change the illumination of each image by adjusting its $Y$ value. In the experiments, we set seven different $\Delta Y$ values and adjust $Y$ by adding $\Delta Y$ to $Y$. Some generated examples are shown in \figref{local_examples2}(a) and \figref{local_examples2}(b). We compare our methods with ResNet50, DenseNet, and PCB. \figref{local_examples2}(c) and \figref{local_examples2}(d) show the results, which demonstrate that our proposed method is more effective for illumination-adaptive datasets synthesizing by varying the $Y$ values.

\section{Conclusion}

This paper raises a new issue, which has not been investigated before as far as we know. Traditional models for the multi-illumination condition may not work well for this task. We propose to disentangle the illumination feature apart to address the new problem. Experimental results illustrate that the traditional model has a significant drop of performance when the illumination of gallery images are different and the scales vary unsteadily, and demonstrate the effectiveness of the proposed network. The idea of addressing the multi-illumination problem can be extended to related video surveillance applications, such as tracking~\cite{cui2013tracking} and activity analysis~\cite{liu2015action2activity,liu2016recognizing,zhang2019cityflow}.

Disentangled representation learning is an effective mechanism to distill meaningful information from the mixed feature representation. As far as we know, we are the first to explore this mechanism to the novel illumination-adaptive person ReID problem. Our illumination-adaptive setting is more practical for long-term ReID. Compared with the work in the field of the face identification, the proposed model further predicts the scale of the illumination and strengthens the re-identification ability of identity features. By showing its effectiveness, we hope the method can also help address problems in other application domains. Also, our main contribution lies in pointing out an important problem for illumination-adaptive person ReID and demonstrating promising initial results. We believe that our paper could inspire more work in this direction and make the ReID techniques more practical.

In the future, we will improve our method in the following aspects: 1) Experiments show that our method does not work very well in very low-light conditions, because of the lose of person-related information. We will investigate how to recover person-related information under such a condition. 2) Although we have used a soft label strategy and adopted the classifier problem to do regression, the framework still tends to disentangle the illumination to different levels of scales. We will study how to estimate the illumination to a continuous value, and make our framework a genuine regression way for the illumination disentanglement. 3) Our method focuses on the illumination problem, while ignoring the other challenges, such as misalignment, occlusion, low-resolution. In the future, we will integrate with other modules to improve ReID performance as a whole.
\small{
\bibliographystyle{ieee}
\bibliography{ieee_ref}
}

\end{document}